\documentclass[english]{article}
\usepackage[T1]{fontenc}
\usepackage[latin9]{inputenc}
\usepackage{booktabs}
\usepackage{babel}
\usepackage{array}
\usepackage{url}
\usepackage{multirow}
\usepackage{algorithm}
\usepackage{algpseudocode}
\usepackage{amsmath}
\usepackage{amsthm}
\usepackage{bbm}
\usepackage{graphicx}
\usepackage{epstopdf}
\usepackage{epsfig}
\usepackage{color}
\usepackage[authoryear]{natbib}
\usepackage{authblk}

\newcommand{\be}{\begin{equation}}
\newcommand{\ee}{\end{equation}}
\newcommand{\bey}{\begin{eqnarray}}
\newcommand{\eey}{\end{eqnarray}}
\newcommand{\beyn}{\begin{eqnarray*}}
\newcommand{\eeyn}{\end{eqnarray*}}

\newcommand{\tr}{\mbox{tr}}
\newcommand{\bmu}{\boldsymbol{\mu}}
\newcommand{\hbeta}{\hat{\beta}}
\newcommand{\halpha}{\hat{\alpha}}
\newcommand{\hd}{\hat{d}}

\newcommand{\calG}{\mathcal{G}}
\newcommand{\calK}{\mathcal{K}}
\newcommand{\calL}{\mathcal{L}}
\newcommand{\calN}{\mathcal{N}}
\newcommand{\calT}{\mathcal{T}}

\newcommand{\E}{\mbox{E}}
\newcommand{\Var}{\mbox{Var}}
\newcommand{\Cov}{\mbox{Cov}}

\newcommand{\ARE}{\operatorname{ARE}}

\newcommand{\btheta}{\boldsymbol{\theta}}
\newcommand{\bphi}{\boldsymbol{\phi}}

\newcommand{\bSigma}{\mathbf{\Sigma}}
\newcommand{\bGamma}{\mathbf{\Gamma}}

\newcommand{\bTheta}{\mathbf{\Theta}}
\newcommand{\barbTheta}{\bar{\bTheta}}
\newcommand{\bPhi}{\boldsymbol{\Phi}}
\newcommand{\barbPhi}{\bar{\bPhi}}

\newcommand{\hbSigma}{\hat{\bSigma}}

\newcommand{\bA}{\mathbf{A}}
\newcommand{\bB}{\mathbf{B}}
\newcommand{\bG}{\mathbf{G}}

\newcommand{\bI}{\mathbf{I}}
\newcommand{\bJ}{\mathbf{J}}
\newcommand{\bV}{\mathbf{V}}

\newcommand{\bb}{\mathbf{b}}

\newcommand{\bu}{\mathbf{u}}

\newcommand{\bz}{\mathbf{z}}

\newcommand{\bzero}{\mathbf{0}}

\newcommand{\bbone}{\mathbbm{1}}

\newcommand{\T}{\top}
\newcommand{\tT}{\tilde{T}}

\newcommand{\dequ}{\stackrel{d}{=}}
\newcommand{\iidsim}{\stackrel{\text{i.i.d.}}{\sim}}
\newcommand{\N}{\calN}

\newtheorem{thm}{Theorem}
\newtheorem{lmm}{Lemma}
\newtheorem{rem}{Remark}

\begin{document}

\title{Group Shapley with Robust Significance Testing and Its Application to Bond Recovery Rate Prediction}

    \author[a]{\small Jingyi Wang}
    \author[a\& b]{\small Ying Chen}
    \author[b]{\small Paolo Giudici}

   \affil[a]{\footnotesize Asian Institute of Digital Finance, National University of Singapore, Singapore}
   \affil[b]{\footnotesize Centre for Quantitative Finance, Department of Mathematics, National University of Singapore, Singapore}
      \affil[c]{\footnotesize Risk Management Institute, National University of Singapore, Singapore}
	\affil[d]{\footnotesize Department of Economics and Management, University of Pavia, Pavia, Italy}



\maketitle

\begin{abstract}
We propose Group Shapley, a metric that extends the classical individual-level Shapley value framework to evaluate the importance of feature groups, addressing the structured nature of predictors commonly found in business and economic data. More importantly, we develop a significance testing procedure based on a three-cumulant chi-square approximation and establish the asymptotic properties of the test statistics for Group Shapley values. Our approach can effectively handle challenging scenarios, including sparse or skewed distributions and small sample sizes, outperforming alternative tests such as the Wald test. Simulations confirm that the proposed test maintains robust empirical size and demonstrates enhanced power under diverse conditions. To illustrate the method's practical relevance in advancing Explainable AI, we apply our framework to bond recovery rate predictions using a global dataset (1996-2023) comprising 2,094 observations and 98 features, grouped into 16 subgroups and five broader categories: bond characteristics, firm fundamentals, industry-specific factors, market-related variables, and macroeconomic indicators. Our results identify the market-related variables group as the most influential. Furthermore, Lorenz curves and Gini indices reveal that Group Shapley assigns feature importance more equitably compared to individual Shapley values.
\end{abstract}


\section{Introduction}\label{sec:intro}
Interpretability has become a cornerstone of modern machine learning (ML), where understanding the rationale behind predictions is as important as achieving high predictive accuracy. The Shapley value, derived from cooperative game theory, has emerged as a widely used metric in Explainable AI (XAI) for quantifying the contribution of individual features to model outputs (see, e.g., \cite{strumbelj2010efficient, vstrumbelj2014explaining, datta2016algorithmic}). However, as feature spaces grow in size and complexity, individual-level Shapley values become overly granular, limiting their interpretability and practical utility. This challenge is particularly pronounced in high-dimensional datasets often encountered in credit risk management, healthcare, and economic policy analysis.

High-dimensional features in real-world datasets can often be grouped into economically meaningful categories, such as firm characteristics or market-related indicators. Grouping features aligns with domain-specific knowledge, enabling more structured and interpretable insights. An important example is bond recovery rate prediction in credit risk management. Unlike metrics such as Exposure at Default (EAD) or Probability of Default (PD), which assess the likelihood and magnitude of credit losses, recovery rates uniquely measure the proportion of funds recoverable after default. Understanding the driven features in predicting recovery rates is critical for risk assessment, portfolio optimization, and pricing strategies, especially for distressed or lower-rated bonds.

Building on the pioneering work of \cite{jullum2021groupshapley}, the Group Shapley value extends the classical Shapley framework by treating predefined feature groups as units of analysis, adhering to the same axioms of fairness, efficiency, and symmetry. While the individual Shapley value is a special case of the Group Shapley value (where each feature constitutes its own group), computing Group Shapley values remains computationally intensive due to the combinatorial nature of coalition formation (\cite{vstrumbelj2014explaining}). Approximation methods such as sampling-based methods  \cite{vstrumbelj2014explaining}, Kernel SHAP (\cite{SHAP2017}) and Tree SHAP (\cite{TreeSHAP2020}) have been developed to address these challenges, leveraging a weighted linear regression or the hierarchical structure of decision trees.

Despite their utility, Shapley values --- whether individual or grouped --- are point estimates and may lack statistical significance, especially in high-dimensional settings with correlated features, skewed distributions, sparse data or small sample size. Existing resampling approaches such as such as bootstrap procedures (\cite{moretti2008combining}, \cite{fryer2020shapley}, \cite{huang2023increasing}) and permutation testing (\cite{roder2021explaining}, \cite{janssen2022application}, \cite{hamilton2023using}) are computationally expensive and often fail to maintain accurate test levels when features exhibit high correlations---a frequent characteristic of high-dimensional economic and financial datasets (\cite{efron1981nonparametric}).

Parametric alternatives like Student's t-test and its extensions also face limitations under non-normal distributions (\cite{Katayama2013}), sparsity, or clustering instead of continuous distributions (\cite{lundberg2017unified}, \cite{ahad2011sensitivity}, \cite{lipsitz1998tests}). For Group Shapley values, the Wald test, based on Hotelling's T$^2$ framework, extends the t-test to multivariate settings. Nevertheless, it is limited by the potential non-invertibility of the sample covariance matrix, particularly when the number of feature groups exceeds the sample size (\cite{BS1996}). The CQ test (\cite{chen2010two}) mitigates this issue by relying on the trace of the covariance matrix. However, both the Wald and CQ tests assume a Gaussian null distribution, an assumption that is frequently violated in real-world datasets, further limiting their applicability (\cite{Katayama2013}, \cite{aoshima2018two}).

Our study seeks to extend the classical individual-level Shapley value framework to Group Shapley, enabling the evaluation of feature group importance and addressing the structured nature of predictors commonly found in business and economic data. To achieve this, we propose Tree Group SHAP, a computationally efficient method that leverages decision tree structures to calculate Group Shapley values. To address the statistical limitations of existing methods, we develop a significance testing procedure based on a three-cumulant chi-square approximation, demonstrating its robust performance across diverse data conditions e.g. sparse or skewed distributions and small sample sizes.

In simulation studies, the proposed test statistic exhibits superior performance across a wide range of scenarios, including normal, symmetric non-normal, and skewed distributions. Under normal conditions, it outperforms existing methods in 24 out of 27 configurations. When data deviate from normality, such as in cases of skewed or sparse distributions, the test maintains robust empirical sizes and demonstrates consistently enhanced power, outperforming all competing methods. Furthermore, this robustness extends to scenarios with varying levels of feature correlation, highlighting the method's versatility and reliability across diverse data conditions.

Our framework is applied to bond recovery rate prediction using a global dataset spanning 1996-2023. The dataset includes 2,094 observations and 98 features grouped into five categories: bond characteristics, firm fundamentals, industry-specific factors, market-related variables, and macroeconomic indicators. By identifying the market-related variables group as the most influential predictor, our results highlight the practical relevance of Group Shapley values for advancing Explainable AI in finance. Furthermore, Lorenz curves and Gini indices reveal that Group Shapley assigns feature importance more equitably compared to individual Shapley values.

Our work builds on existing methodologies while introducing several important distinctions. 1) Extending the foundational Tree SHAP algorithm proposed by \cite{TreeSHAP2020}, we adapt this efficient framework to grouped features, enabling the analysis of structured predictors often encountered in business and economic datasets. This extension achieves computational gains compared to prior methods such as the exact computation approach described in \cite{jullum2021groupshapley}, which can be computationally prohibitive for high-dimensional data. 2)
A central contribution of our study is the development of an asymptotic theory and a significance testing procedure for Group Shapley values. By employing the three-cumulant chi-square approximation method introduced in \cite{Zhang2005}, we approximate the null distribution of Group Shapley test statistics. This approach allows for robust testing under diverse scenarios, including cases with skewed or sparse data and small sample sizes, and provides competitive performance compared to alternative methods such as the Wald and CQ tests. While these alternative tests rely on assumptions such as normality or the invertibility of covariance matrices, our method is designed to handle common challenges found in real-world data, such as feature correlation and non-normal distributions. 3)Previous studies, including \cite{mora2006sovereign}, \cite{jankowitsch2014determinants}, and \cite{NAZEMI2018664}, have identified a wide range of variables affecting recovery rates, encompassing bond characteristics, firm fundamentals, industry-specific factors, market-related variables, and macroeconomic conditions. However, these studies typically examine feature importance at an individual level and do not fully address the benefits of structured group analysis. Additionally, work by \cite{BELLOTTI2021428} compared machine learning models for recovery rate prediction, highlighting the strong performance of ensemble methods such as boosted trees and random forests. While these approaches excel in prediction accuracy, they often lack a systematic way to interpret grouped feature contributions.

In contrast, our framework combines computational efficiency and statistical rigor, enabling the analysis of grouped predictors while ensuring reliable significance testing. This approach provides insights into the contributions of feature groups, complementing predictive accuracy with improved interpretability. By focusing on both group-level importance and robust statistical evaluation, our work addresses a gap in existing methodologies and provides a practical tool for applications such as credit risk management and financial decision-making. Although this study focuses on recovery rate prediction, the methodology is generalizable and holds potential for broader application across domains where understanding group-level feature contributions is critical for informed decision-making, representing a meaningful advancement in Explainable AI. 

The remainder of the paper is organized as follows: Section~\ref{sec:data} describes the dataset and its characteristics. Section~\ref{sec:meds} presents the formulation of Group Shapley values and details the proposed significance testing procedure. Section~\ref{sec:sim} shows simulation studies that evaluate the effectiveness of our methods across various data conditions. Section~\ref{sec:app} demonstrates the application of our framework to bond recovery rate prediction, providing practical insights and results. Finally, Section~\ref{sec:rem} concludes with a discussion of the implications of our findings and potential avenues for future research. Technical proofs and supplementary details are included in the Appendix.

\section{Data}\label{sec:data}

We consider a dataset comprising 98 features and 2094 observations within the period from 1996 to 2023, covering 576 firms and 1638 bonds globally. We have derived the data from four distinct sources. Macro-economic and market related features are obtained from FRED and Refinitiv; financial statement features of firms are sourced from Bloomberg (BBG);  bond-level features, as well as firm-level or market-level credit product features are provided by the Credit Research Initiative (CRI) of the National University of Singapore. 

Recovery rates are defined as the bond's price seven days after the default date:
    \be\label{RR}
        Recovery\; Rate\;(RR) = \frac{Recovered\; Amount}{Face\; Value\; of\; the\; Bond},
    \ee
where the recovered amount may arise from mechanisms like asset liquidation or debt restructuring. During this 7-day period, bond recovery rates can exceed one under certain scenarios, such as the time value of money due to inflation, asset appreciation, overcompensation in debt restructuring, or currency fluctuations. 

\begin{figure}[!ht]
    \centering
    \includegraphics[width=0.6\textwidth]{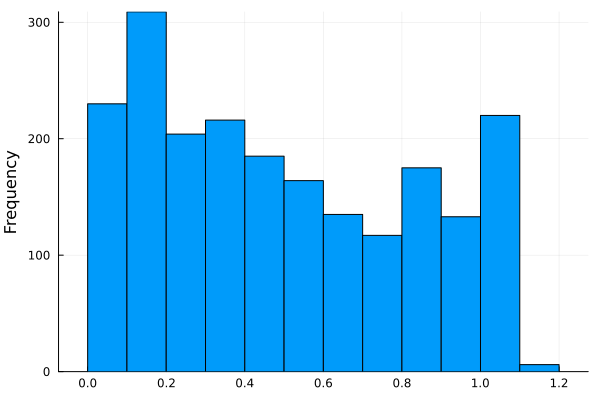}
    \caption{Recovery rates of UP5 from 1996 to 2023, covering 576 firms and 1638 bonds globally}
    \label{fig:RR7}
\end{figure}

Following this definition, the histogram  of the observed recovery rates  from the UP5 dataset (1996-2023), covering 576 firms and 1,638 bonds globally, is displayed in Figure~\ref{fig:RR7}. In the Figure, the  y-axis represents the frequency of each bar, whereas the  x-axis the recovery rate.

Figure~\ref{fig:RR7} reveals a broad range of recovery rates. They are almost all distributed between 0 and 1 and occasionally exceed 1.  Although recovery rates tend to cluster around their mean value, they exhibit significant variability, with a large standard deviation. This wide distribution highlights the limitations of assuming a fixed  recovery rate (such as 40$\%$) in pricing models, which oversimplifies the complex and dynamic nature of actual recovery rates. Assuming a fixed recovery rate  lead to inaccurate risk assessments, flawed pricing strategies, and miscalculated credit risk metrics. 

To complement Figure~\ref{fig:RR7}, summary statistics are presented in Table~\ref{tab:RR7}. 

\begin{table}[!ht]
\centering
\caption{Summary statistics for recovery rates}
\begin{tabular}{cccccccc}
\toprule
$S$ (sample size) & Mean & Std. & Min & 25\% & Median & 75\% & Max \\
\midrule
2094 & 0.4920 & 0.3293 & 0.0000 & 0.1889 & 0.4379 & 0.7966 & 1.1326 \\
\bottomrule
\end{tabular}%
\label{tab:RR7}%
\end{table}%

Both Figure~\ref{fig:RR7} and the standard deviation in Table~\ref{tab:RR7}, equal to about 0.33, highlight a strong variability in the observed recovery rates. Additionally, with a mean of 0.49 exceeding the median of 0.44, the distribution appears right-skewed, and the maximum recovery rate exceeds 1, further illustrating the heterogeneity in recovery outcomes.

A model with 98 features is quite difficult to interpret. It has been documented that recovery rates are influenced by a wide range of factors \cite{JR2014determinant, cochrane2005bond, altman1996almost}, from bond characteristics, firm fundamentals, industry-specific conditions, market dynamics, to macroeconomic indicators. To improve interpretability, we have employed the domain knowledge from the literature on credit recovery, and  aggregated the 98 features into 16 subgroups and, further, into 5 broader groups based on their economic relevance. The obtained groups, along with their the associated meanings are listed in Table~\ref{tab:G16Name}. 

\begin{table}[htbp]
  \centering
  \caption{5 feature groups, 16 feature subgroups and the associated meanings}
    \begin{tabular}[t]{llp{14.75em}}
    \toprule
    \multicolumn{1}{p{4.19em}}{Group} & SubGroup & Description \\
    \midrule
    Bond & BondProperty & bond properties including coupon rates, duration and seniority \\
\cmidrule{2-3}      & Currency & dummy variable for bond currency \\
    \midrule
    Firm & CompanyDTD & companies' DTD, 12-month moving average of DTD and DTD$\_$sigma \\
\cmidrule{2-3}      & CompanyPD$\_$L & long term (> 3yrs) companies' PD \\
\cmidrule{2-3}      & CompanyPD$\_$M & mediun term (7-36 mons) companies' PD \\
\cmidrule{2-3}      & CompanyPD$\_$S & short term companies' PD (< 7mons) \\
\cmidrule{2-3}      & DefaultEvent & occurance of issuers' default events and the event type \\
\cmidrule{2-3}      & Domicile & dummy variable for domicile countries \\
\cmidrule{2-3}      & FinancialStatement & items form financial statements including ROA, current ratio, debt to equity, liquidity ratio and NI over TA \\
\cmidrule{2-3}      & ShareProperty & stocks' performance including marketcap, last price and idiosyncratic risk \\
    \midrule
    \multicolumn{1}{p{4.19em}}{Industry} & Industry & dummy varibles for industry classification from BBG \\
    \midrule
    Market & MarketDTD & aggregated DTD \\
\cmidrule{2-3}      & MarketPD & aggregated PD, aggregation by economies, by industries, or by cross of economies and industries \\
\cmidrule{2-3}      & StockMarket & stock market performance including SP500, SP500/its 1-year MA, stock index return, median marketcap by economy and equity market volatility \\
    \midrule
    Economy & MacroEconomy & macro economic indexes \\
    \cmidrule{2-3} 
      & YieldSpread & US bond yields spreads for different maturities \\
    \bottomrule
    \end{tabular}%
  \label{tab:G16Name}%
\end{table}%

We remark that the classification in Table~\ref{tab:G16Name} derives  from prior knowledge and, as such, is similar to other classifications attempted in the literature.
For example, \cite{altman1996almost} investigated recovery rates on defaulted bonds and emphasized the importance of bond properties. \cite{JR2014determinant} identified key determinants of recovery rates in the U.S. corporate bond market, highlighting the influence of bond characteristics, firm fundamentals, and macroeconomic variables. \cite{cochrane2005bond} analyzed the term structure and the significance of yield spreads. These studies motivates the groups "BondProperty", "YieldSpread", "FinancialStatement", and "MacroEconomy". Some studies focused on the relations between credit risk and recovery rates, like \cite{altman2005link} and \cite{Bruche2010rrpd}, which leads to the groups "CompanyDTD","CompanyPD","MarketDTD" and "MarketPD". 
For the various term structures, we follow the market convention to split "CompanyPD" into short-run, median-run and long-run, represented by "CompanyPD$\_$S", "CompanyPD$\_$M" and "CompanyPD$\_$L" respectively. Here, "short-run" denotes prediction horizons below 6 months (\cite{slrun1998}), while "long-run" denotes prediction horizons over 3 years (\cite{longrun1995}). Additionally, \cite{das2009implied} studied the credit market and the equity market using data including stock prices, stock volatilities and stock market indexes, which supports the groups "ShareProperty" and "StockMarket". The group "DefaultEvent" originates from a report by Moody's, \cite{cantor2008default}, which provided a comprehensive statistical review of default and recovery rates of corporate bond issuers, categorizing default events and their impact on recovery outcomes.

\begin{figure}[htbp]
    \centering
    \includegraphics[width=1.1\textwidth]{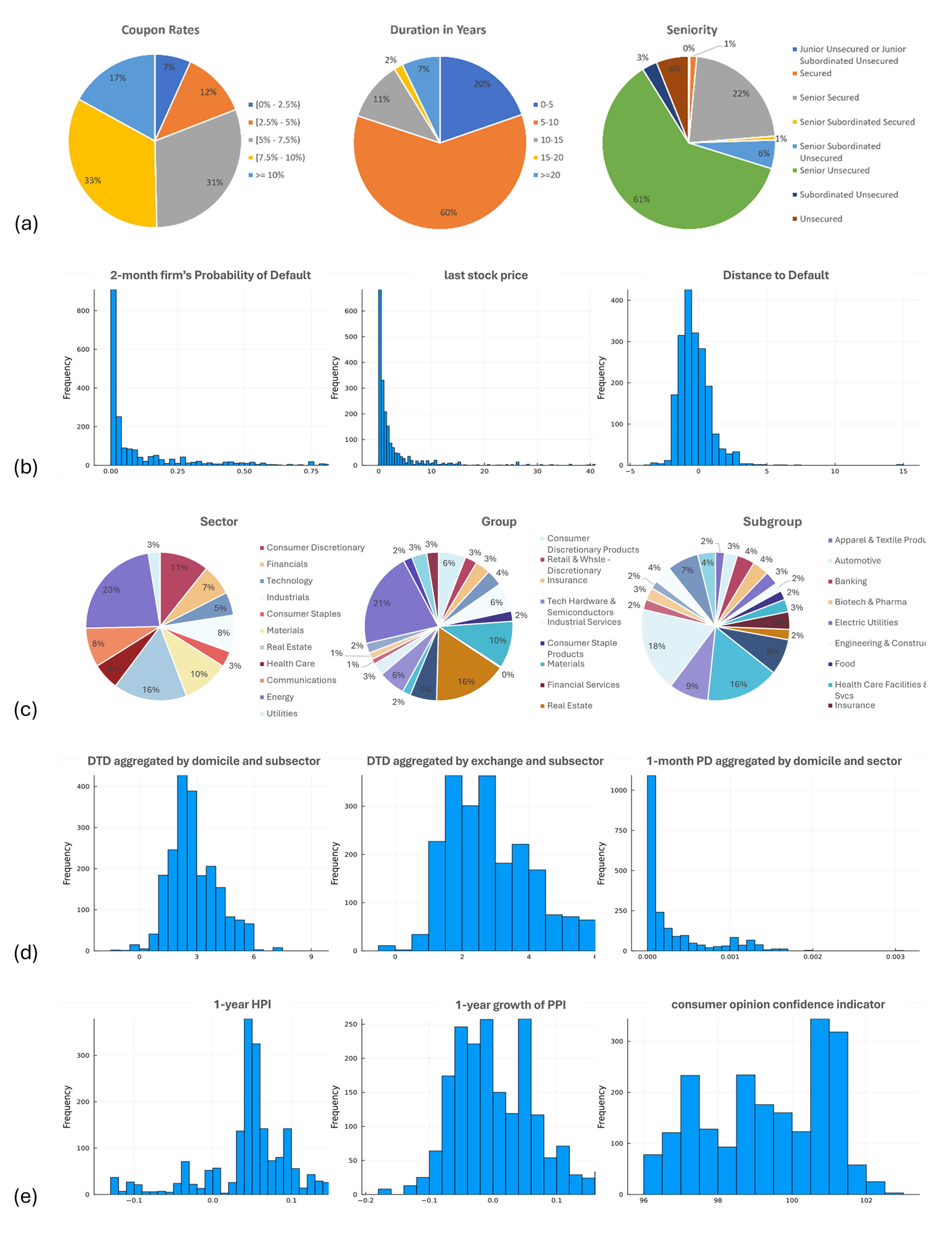}
    \caption{Some explanatory features of UP5, representing  (from top to down): (a)bond characteristics, (b)firm fundamentals, (c)industry-specific factors, (d)market related variables and (e)macroeconomic conditions respectively}
    \label{fig:features}
\end{figure}

We now consider the explanatory features, which will be employed to build a predictive model for the recovery rates. Figure~\ref{fig:features} presents pie charts and histograms illustrating some selected features, categorized as follows (from top to bottom): (a) bond characteristics, (b) firm fundamentals, (c) industry-specific factors, (d) market-related variables, and (e) macroeconomic conditions. These visualizations 
emphasize a complex data structure, which integrates both categorical and numerical features, with different unit of measurement, magnitudes and distribution.

Figure~\ref{fig:features} (a) describes bond characteristics which include coupon rates, duration in years, and seniority, represented by categorical data in pie charts. These features capture diverse attributes such as varying bond durations and seniority levels, indicating their potential role in recovery rates. Part (b) describes firm Fundamentals, displaying numerical features such as PD (probability of default), last price, and DTD (distance-to-default), which are highly skewed, and have  substantial variation in their ranges. Industry-specific factors in (c) comprise categorical data on sectors, groups, and subgroups, emphasizing the heterogeneity across industries. Market-related variables in (d) includes numerical data, such as domicile- and exchange-based metrics, which exhibit clustered distributions and overlapping interdependencies, further complicating the analysis. Part (e) illustrates macroeconomic conditions through numerical features like 1-year HPI (housing returns), PPI growth, and consumer confidence indicators, capturing the dynamic nature of macroeconomic influences on recovery rates. 

The  presented features underline the complexity of the dataset which is obviously greater when all features are considered. Many features exhibit potential correlations, introducing interdependencies that increase the modeling challenge. Additionally, the inclusion of multiple categorical features, such as industry subgroups, creates high-dimensional categorical spaces that interact with numerical variables in a non-linear manner. The  heterogeneity and interdependence among features requires advanced techniques to interpret their joint contributions to recovery rates.



\begin{figure}[ht]
    \centering
    \includegraphics[width=0.8\textwidth]{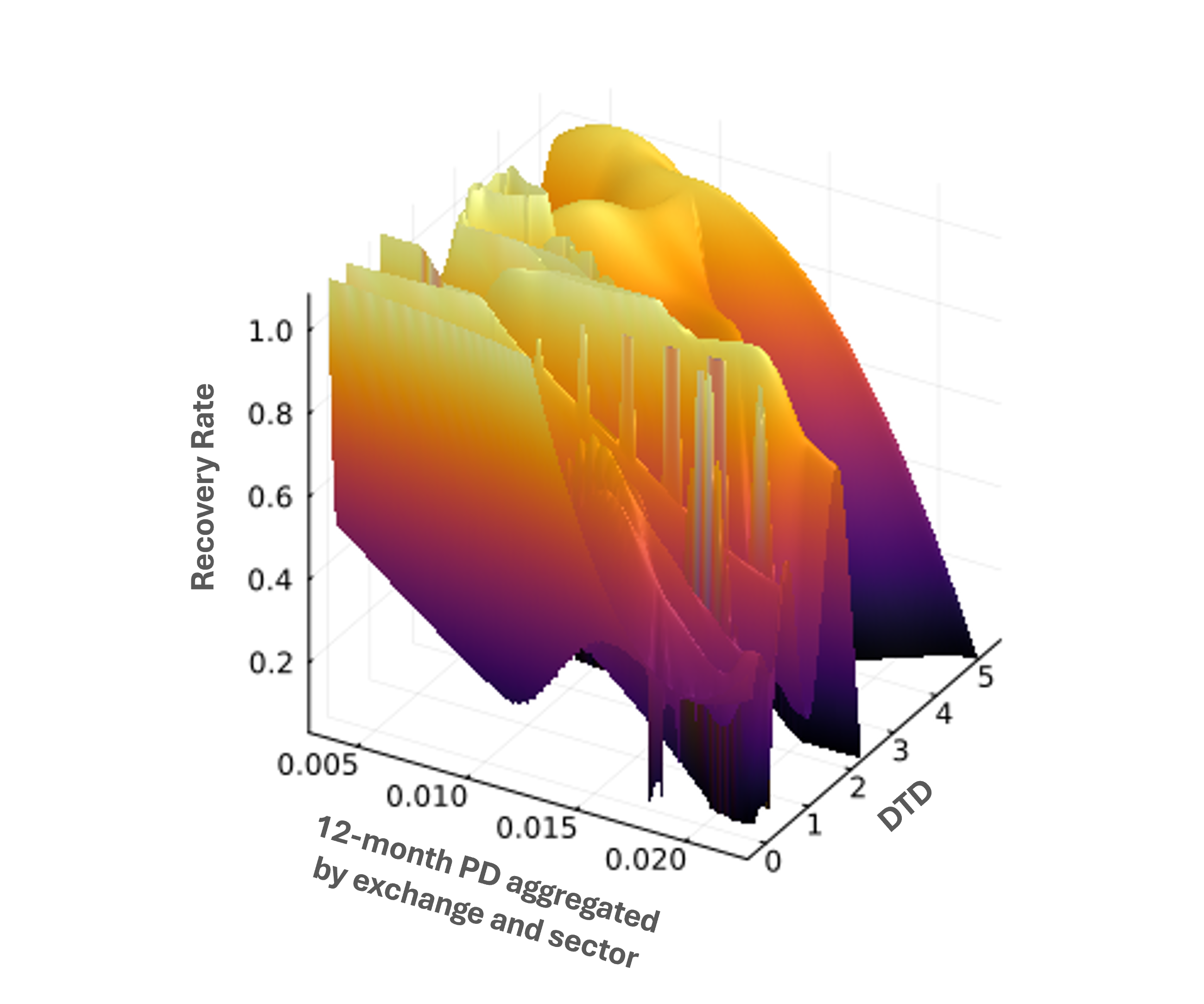}
    \caption{Recovery rates v.s. 12-month exchange-sector aggregated PD and DTD}
    \label{fig:surface}
\end{figure}

To exempolify these arguments, Figure~\ref{fig:surface} illustrates the relationships between recovery rates and two features: 12-month PD aggregated by exchange and sector (x-axis), and DTD (y-axis). 

The surface plot in Figure~\ref{fig:surface} reveals a complex and non-linear relationships between the features and the recovery rate, with no clear trends or linear patterns evident. Such complexity further highlights the challenges in interpreting these relationships directly and underscores the need for advanced techniques to disentangle  the intricate dependencies driving recovery rates. Furthermore, the relationships between recovery rates and individual features are often too granular to provide actionable insights. Aggregating features into a smaller number of groups with meaningful economic interpretations offers a more practical and interpretable approach to understand the drivers of recovery rates.

\section{Methodology}\label{sec:meds}

\subsection{Group Shapley}\label{sec:GS}

Suppose that the available feature variables can  be partitioned in $K$  mutually exclusive groups, with $j =1, \, \ldots, K.$ Then, for the group $g_j$, we have the group Shapley values $\bphi_{g_j}=(\phi_{g_j1},\phi_{g_j2},\ldots,\phi_{g_jS})^T,$ where $S$ is the sample size.

For a group of features $G \subseteq F$, where $F$ represents the complete feature set, the Group Shapley value quantifies the marginal contribution of $G$ by considering its impact across all possible subsets $M \subseteq F \setminus G$. The Group Shapley value $\phi_G$ is computed as
\be\label{GS}
\phi_G = \sum_{M \subseteq F \setminus G} \frac{|M|! \cdot (|F| - |M| - |G|)!}{|F|!} \cdot \Big[ v(M \cup G) - v(M) \Big],
\ee
where $v(M)$ denotes the prediction value when only features in $M$ are included in the model. $|F|$ and $|G|$ represent total number of features and number of features in group $G$ respectively, while $|M|!$ means factorial of the size of subset $M$. 

Note that the above definition contains the individual Shapley values as a special case. Specifically, when each group consists of a single feature (i.e., $K = |F|$), the Group Shapley values equivalent to the individual Shapley values. Thus, the framework seamlessly accommodates both grouped and individual feature contributions. Importantly, in our framework, the Group Shapley value is computed as a single univariate score for each feature group, reducing the dimensionality of feature contributions from $|F|$ (individual features) to $K$ (groups). To illustrate, suppose the $j$-th group contains $j_K$ features. For each feature, the associated individual Shapley values are given by $\bphi_{g_{j_k}} = (\phi_{g_{j_k}1}, \phi_{g_{j_k}2}, \ldots, \phi_{g_{j_k}S})^T$, where $j = 1, \ldots, K$ and $k = 1, \ldots, j_K$. The group Shapley value $\bphi_{g_j}$ can then be regarded as a reduced representation of the individual Shapley values $(\phi_{g_{j_1}}, \phi_{g_{j_2}}, \ldots, \phi_{g_{j_{j_K}}})$, summarizing the contributions of these $j_K$ features within the group into a single, unified measure.

This dimensionality reduction not only enhances computational efficiency but also simplifies the interpretation of feature importance at the group level.


%

\subsection{Significance tests for Group Shapley Values}\label{sec:stats}






Recall the group Shapley values defined above. 
The  significance testing problem we are considering can be expressed either as
\be\label{onetest}
H_0: \E(\bphi_{g_j})=0,
\ee
testing the group Shapley values directly or as
\be\label{jointtest}
    H_0: \E[(\bphi_{g_{j_1}},\ldots,\bphi_{g_{j_{j_K}}})]=\bzero,
\ee
testing all individual Shapley values of the group.

The identified limitations listed in Section~\ref{sec:intro} motivate the development of a novel test statistics that is robust across diverse scenarios, regardless of whether the data is normal or skewed, dense or sparse, and irrespective of the degree of correlation among features. Such statistics should be capable of handling high-dimensional data without relying on the invertibility of covariance matrices, ensuring broad applicability in complex and heterogeneous datasets. To ensure statistical rigor, the metric also need include a derivable asymptotic distribution for reliable inference.

Our proposed  test statistic for group SHAP is inspired by \cite{fan2015power} and  consists of two parts,  defined as follows: $$T_{GS}=T_0+T_1,$$ where $T_1$ is the pivotal for the testing problem and $T_0$ is a data-dependent component aimed at power enhancement, especially for sparse data. 

In previous studies, such as  \cite{fan2015power} and \cite{yu2024power},  the classical Wald test has been employed as $T_1$:
\be\label{Wald}
T_1 = T_{Wald} = \frac{\barbPhi^T\hbSigma^{-1}\barbPhi}{\sqrt{S}}
\ee
where $\barbPhi$ and $\hbSigma$ are the usual sample mean and sample covariance, defined as



\[
\barbPhi=S^{-1}(\sum_{s=1}^{S}\phi_{g_1s},\ldots,\sum_{s=1}^{S}\phi_{g_Ks})^T
\]
and
\be\label{hbSigma}
\hbSigma=(S-1)^{-1}(\bPhi-\bJ_{S\times 1}\barbPhi^{(A)^T})^T(\bPhi-\bJ_{S\times 1}\barbPhi^{(A)^T}).
\ee
Here, $\bJ_{p\times q}$ denotes a $p \times q$ matrix of all ones. However, the associated  sample covariance matrices are often not invertible, when the sample size is smaller than the dimension of features (\cite{BS1996}). Noting that trace provides a scalar measure of the total variability in the data, capturing the aggregate variance across all dimensions, without requiring matrix inversion. Therefore, we replace the inverse of the covariance with the trace:
\be\label{T1}
T_1 = ||\barbPhi||^2-\tr(\hbSigma).
\ee
To derive the null distribution of $T_1$ in (\ref{T1}), let
\be\label{theta}
\bTheta = \bPhi - \bmu
\ee
denote the centralized group SHAP values, and let $\E(\bTheta)=\bzero$ and $\Cov(\bTheta)=\bSigma$. With some simple algebra, we can express $T_{1}$ as
\be\label{T1decomp}
T_{1}=T_{1,0}+2S_{1}+\tr\{\bSigma^2\},
\ee
where
\be\label{T10}
T_{1,0}=\|\barbTheta\|^2-\tr(\hbSigma), \;\mbox{ and } \; S_{1}=(\barbTheta)^{\T}\{\bmu\},
\ee
with $\barbTheta$ the usual sample mean vectors of the centralized group SHAP values for the dataset. $T_{1,0}$ has the same distribution as that of $T_{1}$ under the null hypothesis, i.e. $\bmu = \bzero$.

Note also that $T_{1,0}$ is a quadratic form of the two centralized group SHAP values (\ref{theta}) with $\calK_1(T_{1,0})=\E(T_{1,0})=0$, with \be\label{K2T10}
\calK_2(T_{1,0})=\Var(T_{1,0})=2\left\{\frac{\tr(\bSigma^{2})}{S(S-1)}\right\},
\ee and 
\be\label{K3T10}
\calK_3(T_{1,0})=\E(T_{1,0}^3)=8\left\{\frac{(S-2)\tr(\bSigma^{3})}{S^2(S-1)^2}\right\}.
\ee


We employ the three-cumulant chi-square approximation method (\cite{Zhang2005}) to approximate the null distribution. The main idea is to match the first three cumulants  (means, variances, and third central moments) of $R$ and null statistic, where $R$ is a random variable with approximation parameters $\beta_0, \beta_1$ and $d$:
\be\label{R}
    R=\beta_0+\beta_1 \chi_{d}^{2}.
\ee
The first three cumulants of $R$ are given by  
\[
    \begin{array}{cl}
         \calK_1(R) &= \beta_0+\beta_1 d,  \\
         \calK_2(R) &= 2\beta_1^2 d, \mbox{ and }\\
         \calK_3(R) &= 8\beta_1^3 d.
    \end{array}
\]

After matching the first three cumulants of $T_{1,0}$ and $R$ defined in (\ref{R}), we have that:
\be\label{betadf}
\beta_0=-\frac{2\calK_2^2(T_{1,0})}{\calK_3(T_{1,0})},\;\;
\beta_1=\frac{\calK_3(T_{1,0})}{4\calK_2(T_{1,0})},\;\;\mbox{and}\;\;
d=\frac{8\calK_2^3(T_{1,0})}{\calK_3^2(T_{1,0})}.
\ee
To obtain the estimated approximation parameters $\hbeta_0$, $\hbeta_1$ and $\hd$, we need to estimate $\calK_2(T_{1,0})$ and $\calK_3(T_{1,0})$ consistently from the data (See details in Section~\ref{sec:appendix}).

Then, with the normalized version of $T_{1,0}$, $\tT_{1,0}=T_{1,0}/\sqrt{\widehat{\calK_2(T_{1,0})}}$, the test can be conducted 
using the approximate critical value $c_{\alpha}=\{\chi_{\hd}^{2}(\alpha)-\hd\}/\sqrt{2\hd}$ where $\chi_{d}^2(\alpha)$ denotes the upper $100\alpha$ percentile of $\chi_{d}^2$, or the approximate $p$-value $\Pr( \chi_{\hd}^{2}\ge \hd+\sqrt{2\hd}\tT_{1,0})$.

Note that $T_0$, based the extreme value theory, enhances the test power and does not affect the asymptotic distribution under the null hypothesis, which can be computed as follows:
\be\label{T0}
T_0 = \sqrt{K}\sum_{i=1}^Kh_i\bbone\{h_i\geq9\delta\},
\ee
where $h_i=vm_i/vv_i$, $vm_i=(\barbPhi_i)^2$, $vv_i=\sigma^{2}_{ii}$ and $\hat{\sigma}^{2}_{ii}$ denotes the element in the $i$th-row and $i$th-column of matrix $\hbSigma$. 
\be\label{delta}
\delta=\log(\log(S))^2\log(K)
\ee 
is the pre-determined power enhanced cut-off value, which follows from the large deviation theory, $\max_i\sqrt{h_i}=O_P(\sqrt{\log(K)})$.

To further study the asymptotic size and power of the test, we impose the following condition:
\begin{itemize}
    \item[C1.] For any given dataset, we assume $\bphi_s=\bmu+\bGamma\bz_s$, where $\bphi_s=(\phi_{g_1s},\phi_{g_2s},$\\$\ldots,\phi_{g_Ks})^T;s=1,\ldots,S$; $\bGamma$ is a $K\times p$ matrix for some $p\geq K$ such that $\bGamma\bGamma^{(A)^T}=\bSigma$, and $\bz_s$ are i.i.d. $p$-vectors with $\E(\bz_s)=\bzero$ and $\Cov(\bz_s)=\bI_p$, the $p\times p$ identity matrix. Furthermore, each $z_{sj};j=1,\ldots,p$ in $\bz_{s}=(z_{s1},\ldots,z_{sj},\ldots,z_{sp})^T$ has finite 8-th moment, $\E(z_{sj}^4)=3+\Delta$ for some constant $\Delta$ and for any positive integers $q$ and $\alpha_\ell$ such that $\sum_{\ell=1}^q\alpha_{\ell}\leq 8$, $\E(z_{s\ell_1}^{\alpha_1},\cdots,z_{s\ell_q}^{\alpha_q})=\E(z_{s\ell_1}^{\alpha_1})\cdots,\E(z_{s\ell_q}^{\alpha_q})$ for any $\ell_1\neq\ell_2\neq\cdots\neq\ell_q$ 
\end{itemize}

This condition, also imposed by \cite{chen2010two}, specifies a factor model for high-dimensional data analysis. We can then proof the following Theorem.

\begin{thm}\label{thm:size}
    Under the null hypothesis (\ref{jointtest}) and Condition C1, as $S,K\to\infty$,
    \[
        P(T_{GS}\geq c_{\alpha})\to\alpha.
    \]
\end{thm}

\begin{rem}\label{rem:power} [Theorem 3.3 in \cite{fan2015power}] Consider two alternative spaces 
\[
\calT_s=\{\bPhi:\max_{1\leq i\leq K}(\barbPhi_i)^2/\sigma^{2}_{ii}>9\delta\}
\]
for the sparse alternative where $\delta$ is defined in (\ref{delta}), and 
\[
    \calT_d=\{\bPhi:\|\bPhi\|^2>C\delta K/S\}
\] 
for the dense alternative,  where $C$ is some constant such that $C>0$. 

Then, for any $\alpha \in (0,1)$, as $S,K\to\infty$,
\[
\inf_{\bPhi\in \calT_s}P(T_0>\sqrt{K}|\bPhi)\to 1, \; \; \inf_{\bPhi\in \calT_d}P(T_1>c_{\alpha}|\bPhi)\to 1
\]
and hence
\[
    \inf_{\bPhi\in \calT_s\cup\calT_d}P(T_{GS}>c_{\alpha}|\bPhi)\to 1.
\]
\end{rem}


\begin{rem}  
    Compared to the tests that lack the power enhancement component, such as those presented in \cite{chen2010two} and \cite{zhangguozhouzhu2020}, Remark~\ref{rem:power} justifies a substantial increment in power once $T_0$ is incorporated. This enhancement is evident as the region in which the test is effective expands $\calT_d$ to $\calT_s\cup\calT_d$. 
\end{rem}

\subsection{A tree algorithm to calculate group Shapley values}\label{sec:GTS}



Motivated by the better performance of tree-based models and by the advantages of the tree SHAP, our group SHAP values are developed from tree SHAP, consistently with the predictions that  can be obtained from an ensemble tree model, such as  XGBoost. The algorithm to compute Group Tree Shapley values is given below.



\noindent\textbf{Input:}
\begin{itemize}
    \item A dataset \( X \) with features divided into groups \( \calG = \{g_1, g_2, \ldots, g_K\} \), where each \( g_j \) is a subset of features.
    \item A trained tree-based model with \( T \) decision trees.
    \item Individual SHAP values for all features (optional, for weighted splits).
\end{itemize}

\noindent\textbf{Output:}
\begin{itemize}
    \item Group SHAP values \( \bphi_{g_j} \) for each feature group \( g_j \in \calG \).
\end{itemize}

\noindent\textbf{Procedure:}

\begin{enumerate}
    \item \textbf{Initialization:}
    \begin{itemize}
        \item Set \( \bphi_{g_j} = 0 \) for all groups \( g_j \in \calG \).
        \item Extract tree structures \( t = 1, \ldots, T \) from the trained model.
    \end{itemize}
    
    \item \textbf{Iterate Over Trees:}
    \begin{itemize}
        \item For each tree \( t \) in the model:
        \begin{itemize}
            \item Retrieve all nodes \( n \in N(t) \).
        \end{itemize}
    \end{itemize}

    \item \textbf{Node-Level Computation:}
    \begin{itemize}
        \item For each node \( n \) in tree \( t \):
        \begin{itemize}
            \item Compute the path probability \( P_{t, n} \), defined as the proportion of samples reaching node \( n \).
            \item Compute \( \Delta C_{g, n} \) for each group \( g \), where:
            \[
            \Delta C_{g, n} = (V_L P_L + V_R P_R) - V_{parent}.
            \]
            \item Here, \( V_L \), \( V_R \) are the values at the left and right child nodes, \( P_L \), \( P_R \) are the proportions of samples reaching the left and right child nodes, and \( V_{parent} \) is the value at the parent node before the split.
        \end{itemize}
    \end{itemize}

    \item \textbf{Group Feature Splits:}
    \begin{itemize}
        \item For nodes splitting on a group \( g \), determine the split using a weighted combination of the features in \( g \):
        \[
        X_{g, \text{split}} = \sum_{f \in g} w_f X_f,
        \]
        where \( w_f \) are weights based on individual SHAP values or pre-defined criteria.
    \end{itemize}

    \item \textbf{Aggregate Contributions:}
    \begin{itemize}
        \item Update the group SHAP value for \( g \):
        \[
        \bphi_{g_j} = \Delta C_{g, n} P_{t, n}.
        \]
    \end{itemize}

    \item \textbf{Repeat for All Groups:}
    \begin{itemize}
        \item Repeat steps 3--5 for all groups \( g_j \in \calG \).
    \end{itemize}

    \item \textbf{Output Results:}
    \begin{itemize}
        \item Return \( \bphi_{g_j} \) for all groups \( g_j \in \calG \).
    \end{itemize}
\end{enumerate}

This algorithm leverages the hierarchical structure of decision trees to efficiently compute group-level contributions. To determine the split point at node $n$ for each $X_g$ we adopt a classical approach, following studies such as \cite{murthy1994system}, \cite{geurts2006extremely}, and \cite{menze2011oblique}, which employs a weighted combination of variables. Here, weights are assigned to each variable based on their individual SHAP values. The weighted sum of these variables is treated as a single, composite feature to determine the split.

\section{Simulation Study}\label{sec:sim}

In this section, we evaluate the significance and statistical power of $T_{GS}$ under various dataset distributions, including normal, non-normal but symmetric, and skewed scenarios, as defined below.

\begin{description}
    \label{samdes}
    \item[Normal Model:] $z_{sj},j=1,\ldots,K \iidsim \N(0,1)$,
    \item[Symmetric Model:] $z_{sj}= u_{sj}/\sqrt{2}$, with $u_{sj},j=1,\ldots,K \iidsim t_4$,
    \item[Skewed Model:] $z_{sj}= (u_{sj}-1)/\sqrt{2}$, with $u_{sj},j=1,\ldots,K \iidsim \chi^2_1$,
\end{description}

with different sample sizes, dimensionalities, and correlation structures. The performance of $T_{GS}$ is compared against several existing methods for the zero-mean vector testing problem (\ref{jointtest}). Benchmarks include the widely used $T_{Wald}$ and $T_{CQ}$ (\cite{chen2010two}). 

\begin{table}[htbp]
  \centering
  \setlength{\tabcolsep}{3pt}	
  \caption{Empirical sizes (in $\%$) of $T_{Wald}$, $T_{CQ}$ and $T_{GS}$ with nominal size of 5$\%$.}
    \begin{tabular}{cccccccccccc}
    \toprule
    \multirow{2}[4]{*}{Model} & \multirow{2}[4]{*}{$K$} & \multirow{2}[4]{*}{$S$} & \multicolumn{3}{c}{$\rho=0.2$} & \multicolumn{3}{c}{$\rho=0.5$} & \multicolumn{3}{c}{$\rho=0.8$} \\
\cmidrule{4-12}      &   &   & $T_{Wald}$ & $T_{CQ}$ & $T_{GS}$ & $T_{Wald}$ & $T_{CQ}$ & $T_{GS}$ & $T_{Wald}$ & $T_{CQ}$ & $T_{GS}$ \\
    \midrule
    \multirow{9}[6]{*}{Normal} & \multirow{3}[2]{*}{20} & 50 & \textbf{5.35} & 6.45 & 5.45 & 6.00 & 7.02 & \textbf{5.35} & 5.60 & 6.89 & \textbf{5.50} \\
      &   & 300 & 5.20 & 6.23 & \textbf{5.05} & 5.34 & 6.19 & \textbf{5.16} & 5.89 & 7.71 & \textbf{5.77} \\
      &   & 600 & 5.13 & 5.96 & \textbf{4.94} & 5.93 & 7.43 & \textbf{5.63} & 5.41 & 6.97 & \textbf{5.31} \\
\cmidrule{2-12}      & \multirow{3}[2]{*}{100} & 50 & NaN & 6.79 & \textbf{5.57} & NaN & 7.32 & \textbf{5.30} & NaN & 7.22 & \textbf{5.10} \\
      &   & 300 & 5.98 & 6.66 & \textbf{5.86} & 5.97 & 7.12 & \textbf{5.04} & \textbf{5.41} & 7.19 & 5.52 \\
      &   & 600 & \textbf{5.79} & 5.90 & 5.82 & 5.80 & 7.06 & \textbf{5.86} & 4.92 & 6.48 & \textbf{4.97} \\
\cmidrule{2-12}      & \multirow{3}[2]{*}{500} & 50 & NaN & 6.66 & \textbf{5.61} & NaN & 7.63 & \textbf{4.89} & NaN & 7.64 & \textbf{5.68} \\
      &   & 300 & NaN & 7.06 & \textbf{5.92} & NaN & 6.68 & \textbf{5.69} & NaN & 6.64 & \textbf{4.82} \\
      &   & 600 & 6.44 & 6.61 & \textbf{5.54} & 6.69 & 6.96 & \textbf{5.39} & 5.17 & 6.68 & \textbf{5.10} \\
    \midrule
    \multirow{9}[6]{*}{Symmetric} & \multirow{3}[2]{*}{20} & 50 & 6.83 & 6.32 & \textbf{5.10} & 6.12 & 6.92 & \textbf{5.92} & 6.61 & 7.32 & \textbf{5.80} \\
      &   & 300 & 6.33 & 6.25 & \textbf{5.31} & 6.56 & 6.55 & \textbf{5.34} & 6.06 & 6.83 & \textbf{5.18} \\
      &   & 600 & 7.01 & 5.96 & \textbf{4.91} & 6.59 & 6.53 & \textbf{5.58} & 6.27 & 6.96 & \textbf{5.32} \\
\cmidrule{2-12}      & \multirow{3}[2]{*}{100} & 50 & NaN & 6.74 & \textbf{5.01} & NaN & 7.36 & \textbf{5.78} & NaN & 7.38 & \textbf{5.26} \\
      &   & 300 & 6.19 & 6.51 & \textbf{5.06} & 6.45 & 6.72 & \textbf{5.46} & 6.56 & 7.52 & \textbf{5.70} \\
      &   & 600 & 6.62 & 6.21 & \textbf{5.63} & 7.14 & 7.70 & \textbf{5.22} & 6.27 & 6.43 & \textbf{5.41} \\
\cmidrule{2-12}      & \multirow{3}[2]{*}{500} & 50 & NaN & 6.78 & \textbf{5.18} & NaN & 7.28 & \textbf{5.00} & NaN & 7.20 & \textbf{5.58} \\
      &   & 300 & NaN & 6.79 & \textbf{5.33} & NaN & 7.07 & \textbf{5.02} & NaN & 6.45 & \textbf{4.95} \\
      &   & 600 & 6.30 & 6.61 & \textbf{5.04} & 7.61 & 7.41 & \textbf{5.83} & 6.47 & 7.07 & \textbf{5.30} \\
    \midrule
    \multirow{9}[6]{*}{Skewed} & \multirow{3}[2]{*}{20} & 50 & 7.06 & 6.05 & \textbf{5.32} & 7.06 & 7.18 & \textbf{5.18} & 7.16 & 7.78 & \textbf{5.42} \\
      &   & 300 & 7.29 & 6.17 & \textbf{5.54} & 6.64 & 6.70 & \textbf{5.47} & 7.26 & 7.39 & \textbf{5.07} \\
      &   & 600 & 6.89 & 6.27 & \textbf{5.13} & 7.34 & 6.80 & \textbf{5.22} & 6.95 & 6.66 & \textbf{4.80} \\
\cmidrule{2-12}      & \multirow{3}[2]{*}{100} & 50 & NaN & 7.09 & \textbf{5.58} & NaN & 7.15 & \textbf{5.08} & NaN & 7.24 & \textbf{5.11} \\
      &   & 300 & 6.28 & 6.54 & \textbf{5.27} & 6.89 & 7.14 & \textbf{5.89} & 7.71 & 7.31 & \textbf{5.45} \\
      &   & 600 & 6.57 & 6.91 & \textbf{5.82} & 7.70 & 6.91 & \textbf{5.82} & 7.35 & 6.58 & \textbf{5.56} \\
\cmidrule{2-12}      & \multirow{3}[2]{*}{500} & 50 & NaN & 7.27 & \textbf{5.45} & NaN & 7.68 & \textbf{5.38} & NaN & 7.04 & \textbf{5.82} \\
      &   & 300 & NaN & 6.98 & \textbf{5.60} & NaN & 7.35 & \textbf{5.09} & NaN & 7.23 & \textbf{5.57} \\
      &   & 600 & 7.21 & 6.77 & \textbf{5.40} & 7.53 & 6.79 & \textbf{5.46} & 7.32 & 6.69 & \textbf{5.43} \\
    \midrule
    \multicolumn{3}{c}{ARE} & NaN & 30.77 & \textbf{7.96} & NaN & 41.22 & \textbf{8.35} & NaN & 41.11 & \textbf{7.72} \\
    \bottomrule
    \end{tabular}%
   \label{tab:size}%
\end{table}%

To carry out this simulation, we maintain a nominal size $\alpha$ of $5\%$ and perform 10,000 simulation iterations. The empirical size or power of each test is then computed based on the fraction of rejections observed among these iterations. Simulated samples are generated from the factor model in Condition C1, $\bphi_{s} =\bmu + \bSigma^{1/2}\bz_{s},s=1,...,S;$ where $\bz_{s}=(z_{s1},\ldots,z_{sK})^\T$ are i.i.d. random variables with $\E(\bz_{s})=\bzero$ and $\Cov(\bz_{s})=\bI_K$. The $K$ entries of $\bz_{s}$ are generated using  the three models aforementioned.

The covariance matrices $\bSigma$ are determined by $\bSigma=\sigma^{2}[(1-\rho)\bI_K+\rho\bJ_K]$. Here, $\bJ_K$ represents the $K\times K$ matrix of ones, where values of $\sigma^{2}$ control the variances of the generated samples, and $\rho$ controls their correlations. We specify $\sigma^2=4$, and $\rho=0.2,0.5$ and 0.9 for nearly uncorrelated, moderately correlated and highly correlated samples respectively. When $\bmu=\bzero$, samples are generated under the null hypothesis in (\ref{jointtest}) for computing the empirical sizes of the tests under consideration. To compare test powers, we $\bmu$ from the following two sets for the $\calT_s$ and $\calT_d$, the sparse and dense alternatives:
    \be\label{sparse}
    \mu_{s} =
        \begin{cases} 
        0.5 & s\leq ceil(K/S)\\ 
        0 & \mbox{otherwise} 
        \end{cases}
    \ee
    and
    \be\label{dense}
    \mu_{s} =
        \begin{cases} 
        \sqrt{\frac{\log(K)}{S}} & s\leq ceil(\sqrt{K})\\ 
        0 & \mbox{otherwise} 
        \end{cases}
    \ee
    respectively, where $ceil()$ means round up.

To evaluate a test's overall performance in maintaining the nominal size, we adopt the average relative error (ARE) metric, as computed by $\ARE=100T^{-1}\sum_{t=1}^T|\halpha_t-\alpha|/\alpha$, where $\halpha_t, t=1,\ldots,T$ represents the empirical sizes across $T$ simulation times. A smaller ARE value for a test implies better overall performance in terms of size control.


\begin{table}[htbp]
  \centering
  \caption{Empirical power (in $\%$) of $T_{Wald}$, $T_{CQ}$ and $T_{GS}$ with $\rho=0.5$ under two alternatives.}
    \begin{tabular}{ccccccccc}
    \toprule
    \multirow{2}[4]{*}{Model} & \multirow{2}[4]{*}{$K$} & \multirow{2}[4]{*}{$S$} & \multicolumn{3}{c}{$\calT_s$} & \multicolumn{3}{c}{$\calT_d$} \\
\cmidrule{4-9}      &   &   & $T_{Wald}$ & $T_{CQ}$ & $T_{GS}$ & $T_{Wald}$ & $T_{CQ}$ & $T_{GS}$ \\
    \midrule
    \multirow{9}[5]{*}{Normal} & \multirow{3}[2]{*}{20} & 50 & 39.77 & 40.07 & \textbf{60.74} & 81.19 & 81.55 & \textbf{82.92} \\
      &   & 300 & 46.47 & 46.86 & \textbf{61.09} & 89.92 & 88.26 & \textbf{90.80} \\
      &   & 600 & 56.10 & 58.96 & \textbf{69.34} & 99.25 & \textbf{99.60} & 98.22 \\
\cmidrule{2-9}      & \multirow{3}[2]{*}{100} & 50 & NaN & 43.27 & \textbf{63.31} & NaN & 84.77 & \textbf{85.06} \\
      &   & 300 & 52.25 & 50.31 & \textbf{68.44} & 93.50 & \textbf{94.50} & 92.86 \\
      &   & 600 & 64.08 & 60.21 & \textbf{72.40} & 96.75 & 96.91 & \textbf{98.26} \\
\cmidrule{2-9}      & \multirow{3}[1]{*}{500} & 50 & NaN & 47.69 & \textbf{76.73} & NaN & 85.44 & \textbf{86.08} \\
      &   & 300 & NaN & 52.03 & \textbf{75.07} & NaN & 89.77 & \textbf{90.14} \\
      &   & 600 & 64.30 & 65.94 & \textbf{82.32} & 90.76 & 92.90 & \textbf{94.24} \\
      \midrule
    \multirow{9}[5]{*}{Symmetric} & \multirow{3}[1]{*}{20} & 50 & 43.47 & 45.38 & \textbf{72.06} & 81.66 & 82.53 & \textbf{82.86} \\
      &   & 300 & 50.12 & 60.30 & \textbf{76.01} & 86.32 & 86.35 & \textbf{86.97} \\
      &   & 600 & 61.18 & 67.47 & \textbf{79.01} & 94.32 & 95.00 & \textbf{96.48} \\
\cmidrule{2-9}      & \multirow{3}[2]{*}{100} & 50 & NaN & 56.74 & \textbf{84.28} & NaN & 87.30 & \textbf{88.53} \\
      &   & 300 & 63.07 & 70.95 & \textbf{83.00} & 89.51 & 89.05 & \textbf{89.60} \\
      &   & 600 & 74.76 & 77.86 & \textbf{87.14} & 93.49 & 94.82 & \textbf{94.86} \\
\cmidrule{2-9}      & \multirow{3}[2]{*}{500} & 50 & NaN & 61.11 & \textbf{87.84} & NaN & 88.23 & \textbf{88.68} \\
      &   & 300 & NaN & 72.33 & \textbf{87.32} & NaN & \textbf{92.31} & 91.74 \\
      &   & 600 & 76.83 & 77.41 & \textbf{83.99} & 94.73 & 94.70 & \textbf{94.95} \\
    \midrule
    \multirow{9}[6]{*}{Skewed} & \multirow{3}[2]{*}{20} & 50 & 62.50 & 62.25 & \textbf{83.56} & 83.27 & \textbf{83.40} & 82.07 \\
      &   & 300 & 70.22 & 78.26 & \textbf{86.73} & 84.04 & 83.91 & \textbf{84.83} \\
      &   & 600 & 81.15 & 79.68 & \textbf{87.05} & \textbf{92.92} & 92.88 & 92.31 \\
\cmidrule{2-9}      & \multirow{3}[2]{*}{100} & 50 & NaN & 63.22 & \textbf{81.00} & NaN & \textbf{91.46} & 91.27 \\
      &   & 300 & 75.85 & 75.47 & \textbf{88.34} & 93.44 & 94.06 & \textbf{94.77} \\
      &   & 600 & 76.39 & 76.62 & \textbf{86.37} & 97.07 & 98.01 & \textbf{98.52} \\
\cmidrule{2-9}      & \multirow{3}[2]{*}{500} & 50 & NaN & 70.73 & \textbf{84.92} & NaN & 89.89 & \textbf{90.45} \\
      &   & 300 & NaN & 82.50 & \textbf{83.16} & NaN & 94.61 & \textbf{95.17} \\
      &   & 600 & 77.04 & 83.24 & \textbf{86.61} & \textbf{96.62} & 94.86 & 95.27 \\
    \bottomrule
    \end{tabular}%
  \label{tab:power}%
\end{table}%

Table~\ref{tab:size} demonstrates the empirical sizes of these three tests under various settings and  provides several interesting findings. For the consistency in size-control, $T_{GS}$ maintains empirical sizes close to the nominal level of 5$\%$ across different models, sample sizes, and correlation structures. $T_{CQ}$ generally shows higher empirical sizes compared to the nominal level, particularly noticeable in high-dimensional settings ($K=500$). $T_{Wald}$ keeps a good performances in size-control for lower dimensions scenarios while fails to produce valid results (NaN) for higher dimensions due to non-invertible covariance matrices, which limits the usage of $T_{Wald}$ especially for our real datasets. The ARE values also indicates the superior performance of $T_{GS}$ in size-control, since $T_{GS}$ has significantly lower ARE values compared to $T_{CQ}$, particularly when the correlation is high $\rho = 0.8$. Comparing the impact of correlations, it is seen that higher correlation tends to increase the empirical sizes for all tests, but $T_{GS}$ remains closest to the nominal level, demonstrating robustness. Overall, $T_{GS}$ demonstrates a robust ability to maintain the nominal size across various scenarios, outperforming the existing methods, especially in high-dimensional settings and under different correlation structures. This indicates its potential as a reliable method for the problem (\ref{jointtest}).

Table~\ref{tab:power} presents the empirical power (in $\%$) of three tests, $T_{Wald}$, $T_{CQ}$, and $T_{GS}$, for the two-sample equal mean vectors testing problem. The power is evaluated under two alternatives, $\calT_s$, the sparse alternative give in (\ref{sparse}), and $\calT_d$, the dense alternative given in (\ref{dense}), with a correlation coefficient $\rho = 0.5$.

From Table~\ref{tab:power}, all tests perform similarly well under the dense alternative ($\calT_d$), achieving high power across different models and sample configurations. However, under the sparse alternative ($\calT_s$), $T_{GS}$ consistently outperforms $T_{Wald}$ and $T_{CQ}$  across all models and sample configurations, which indicates that $T_{GS}$ is more sensitive in detecting differences when the signal is sparse. Overall, $T_{GS}$ shows superior performance in scenarios with sparse signals and comparable performance under dense alternatives, making it a preferred choice for such testing problems.

\section{Recovery Rate Prediction}\label{sec:app}

In this Section we apply our proposed methodology to the described data. 
We employ Gradient Boosting (XGBoost) to predict recovery rates, and to compute group tree SHAP values  according to the algorithm detailed in Section~\ref{sec:GTS}. Subsequently, we apply the test statistic proposed in Section~\ref{sec:stats}.

\begin{figure}[http]
    \centering
    \includegraphics[width=0.9\textwidth]{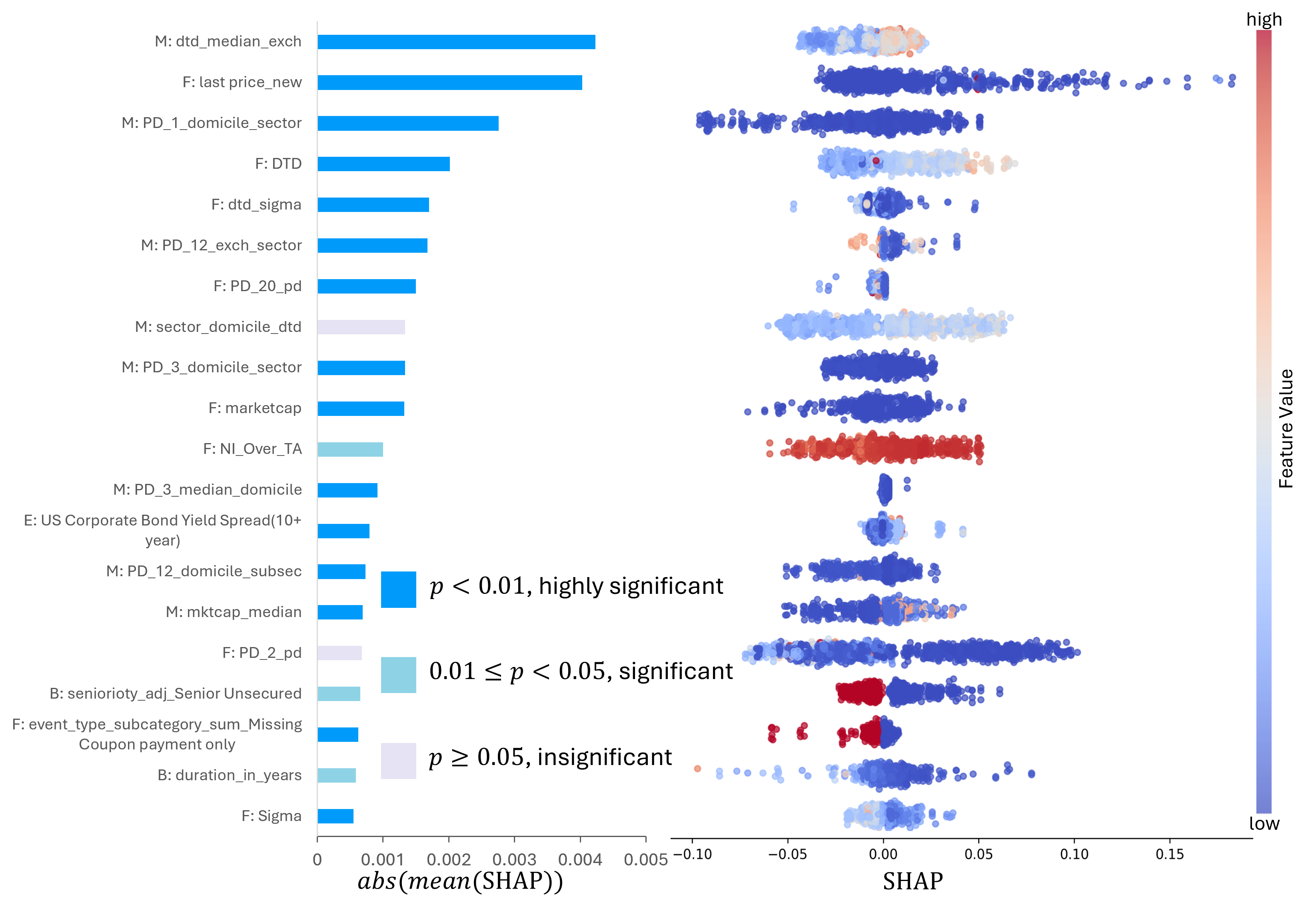}
    \caption{SHAP values for the top 20 individual features, among the 98 available ones.}
    \label{fig:SHAP20}
\end{figure}

Figure~\ref{fig:SHAP20}  presents a dual visualization of the top 20 individual features ranked by the absolute mean of their SHAP values. The left panel showcases a bar chart representing the overall contribution of each feature to the model predictions, with distinct colors indicating varying levels of statistical significance. This highlights not only the magnitude of contributions but also their reliability. The right panel provides a detailed distribution of SHAP values for each feature, illustrating how the magnitude and direction of feature values impact the model's output. 

The left panel of Figure~\ref{fig:SHAP20} shows that "dtd$\_$median$\_$exch", denoting the DTD aggregated by exchanges, has the highest mean absolute SHAP values. It is thus a key feature contribution to the model's predictions. However, a closer inspection of the right panel reveals that the SHAP values for "PD$\_$2$\_$pd" are symmetrically distributed around 0, with high density near the center. This distribution raises questions about its statistical significance, suggesting that its large contribution magnitude might not reliably reflect its true importance. This highlights the necessity of rigorous significance testing to ensure that high-contribution features are genuinely impactful.

It is not surprising to find that even with help of Shapley values, a model with 98 features is quite difficult to  interpret, highlighting the importance of grouping.  

\begin{figure}[http]
    \centering
    \includegraphics[width=0.95\textwidth]{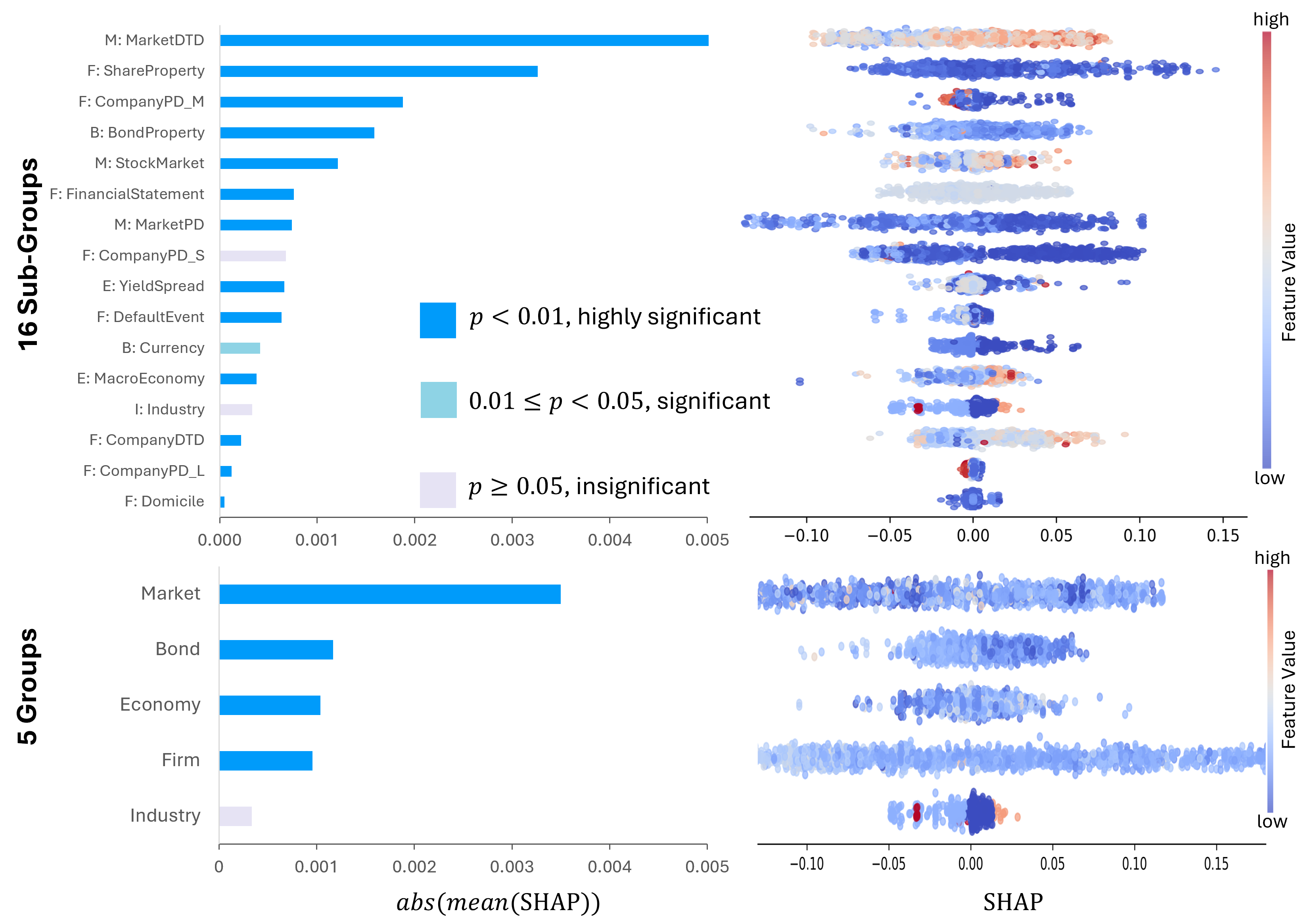}
    \caption{Group SHAP values for 16 subgroups (top) and 5 groups (bottom)}
    \label{fig:MAGS}
\end{figure}

Having simplified the significant 98 features into 16 subgroups and 5 groups, as described in Table~\ref{tab:G16Name},   we can have a more direct insights into the feature contributions. Figure~\ref{fig:MAGS} displays the mean of the absolute Shapley (SHAP) values for the 16 subgroups in the left and 5 groups in the right. Different colors represent different significance levels.

Comparing the mean Group Shapley values in Figure~\ref{fig:MAGS} to the individual Shapley values in Figure~\ref{fig:SHAP20}, note that
while the latter provide detailed attributions for each feature, the high dimensionality and overlapping contributions make their  interpretation, and the individuation of an overall pattern, quite challenging. For instance, interpreting the importance of "dtd$\_$median$\_$exch" or "PD$\_$1$\_$domicile$\_$sector" in isolation may obscure their collective influence as related variables. 
In contrast, the group SHAP values in Figure~\ref{fig:MAGS} aggregate related features (such as  "Market$\_$DTD" and "ShareProperty") into economically meaningful categories, simplifying the interpretation. This aggregated view highlights the dominant impact of macro-level factors like "MarketDTD" and "StockMarket," enabling clearer insights into the underlying drivers of the model's predictions. 



In more detail, from Figure~\ref{fig:MAGS} note that the "Market$\_$DTD" group, representing aggregated DTD across diversified markets, contributes the most on the model's predictions. This suggests that aggregated DTD is a crucial factor for recovery rate. The second and third most important groups, are  "ShareProperty" and "CompanyPD$\_$M",  indicating that they also play a vital role in influencing the recovery rate. Groups like "Industry" and "domicile" have the least impact, suggesting that broader industry categorizations and domicile information are less relevant to the model's predictions compared to more specific financial and credit-related features. In the bottom panel, features are aggregated into 5 broader groups (e.g., Firm, Market, Bond, Macroeconomy, Industry), simplifying interpretation by reducing dimensionality. The Market group is the most impactful, followed by Bond and Macro groups, while Firm and Industry groups contribute minimally, which is consistent with results of 16 subgroup SHAP values. 

Before drawing final conclusions, it is crucial to test the significance of the results. To this end, we applied our proposed test in (\ref{jointtest}) to evaluate whether the group Shapley values are statistically significant, with the results presented in Figure~\ref{fig:MAGS}. Notably, both (\ref{onetest}) and (\ref{jointtest}) are applicable for assessing the significance of group Shapley values within our framework. However, the reduced form associated with test (\ref{onetest}) increases variance, which can lead to an underestimation of the significance level. 

For instance, in the case of the "FinancialStatement" subgroup (among the 16 subgroups) and "Firm" (within the 5 broader groups), these would be deemed "insignificant" if tested using (\ref{onetest}). Upon closer examination of their respective distributions in the right panel of Figure~\ref{fig:MAGS}, both exhibit symmetric shapes with wide ranges, which contribute to their lower significance levels when evaluated simply as a whole. This underscores the limitations of the reduced-form approach in (\ref{onetest}) and highlights the necessity of the joint test (\ref{jointtest}), particularly for robust analysis in high-dimensional datasets. Furthermore, it emphasizes the importance of a statistically rigorous test to ensure accurate interpretations of group-level Shapley contributions.

In the Figure,  the  significant groups are in bright colors while the less significant groups in subdued colors. It is quite interesting to see that the magnitudes do not  always align with the significance level. For example, "CompanyPD$\_$S" ranks 8 among the 16 groups representation but poses a low statistical significance, which may lead to overfitting or misinterpretation. The apparent misalignment between the feature magnitude (expressed by the absolute SHAP value) and its significance level underscores the need for rigorous statistical validation. This prevents over-reliance on features that contribute heavily but lack meaningful predictive power.

From the economic viewpoint, our   analysis considering both SHAP magnitude and statistical significance reveals that market-related variables and bond characteristics are pivotal to predicting recovery rates, followed by macroeconomic conditions and firm fundamentals, which also show notable influence. On the other hand, industry-specific factors are the least impactful in driving recovery rate predictions.

We finally compare the individual and the group based Shapley values in terms of two important statistical aspects: their concentration and their correlation.

\begin{figure}[http]
    \centering
    \includegraphics[width=1.1\textwidth]{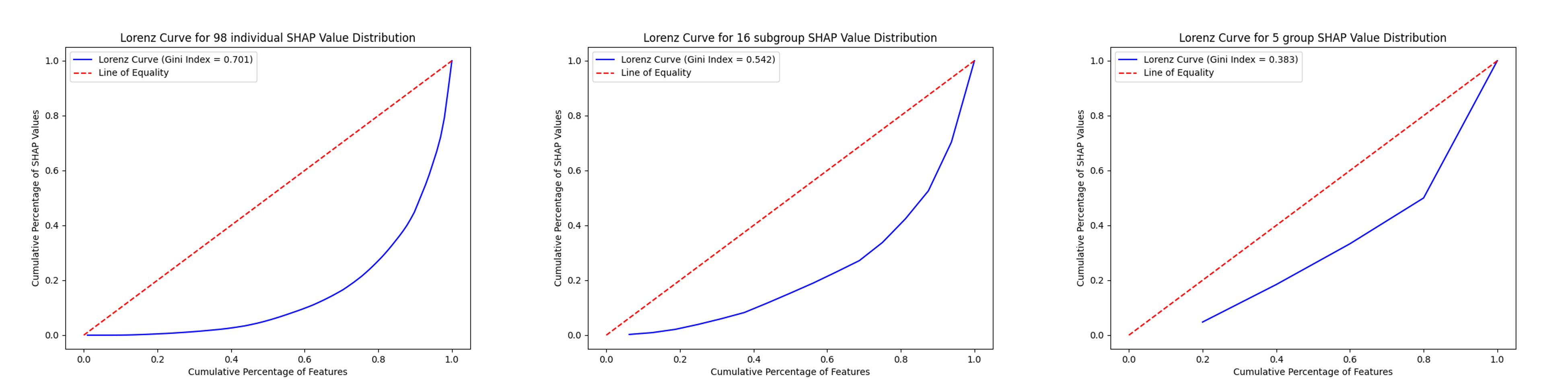}
    \caption{Lorenz curves with Gini index for individual SHAP (left), 16 subgroup SHAP (center) and 5 group SHAP (right)}
    \label{fig:concen}
\end{figure}

Concerning concentration, Figure~\ref{fig:concen} represents the the Lorenz curves with Gini index for both Shapley, 16 subgroup and 5 group Shapley explanations, after we normalise (divide by their sum) and  order them from the lowest  to the highest normalised value.

 The high Gini index of 0.701 in the left panel of Figure~\ref{fig:concen} demonstrates that a few features dominate the contribution, potentially overshadowing less significant features. This can make the interpretation of individual SHAP values overwhelming and misleading in high-dimensional datasets. Aggregating features into meaningful groups reduces inequality in contributions, as evidenced by the lower Gini index of 0.542 in the center and 0.383 in the right panel. This highlights the effectiveness of grouping in simplifying the interpretation and making the contributions more economically meaningful.



\begin{figure}[http]
    \centering
\includegraphics[width=1.1\textwidth]{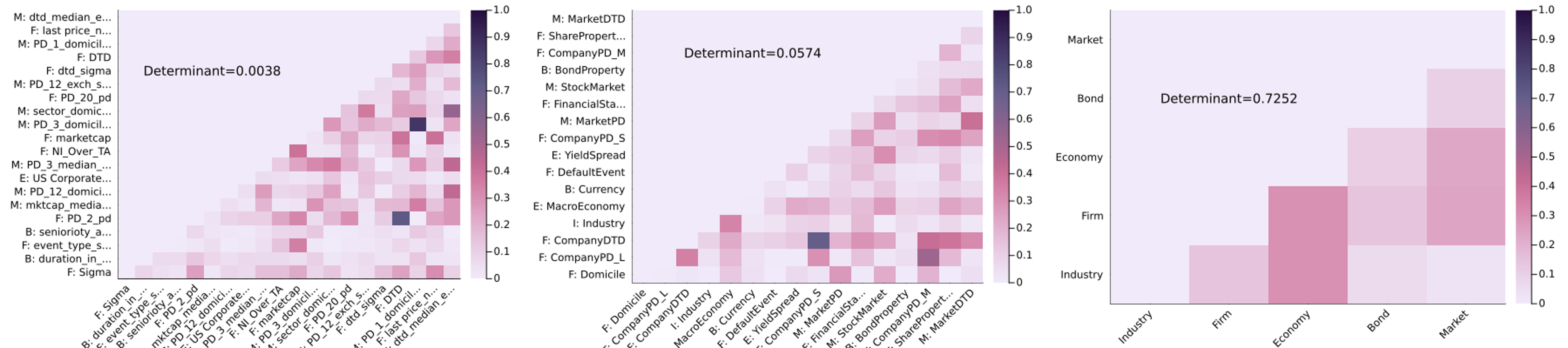}
    \caption{Absolute values of correlation maps for the top 20 individual SHAP (left), 16 subgroup SHAP (center) and 5 group SHAP (right)}
    \label{fig:corrg}
\end{figure}

Concerning correlation, Figure~\ref{fig:corrg}  represents the correlation map among the 20 most important individual features in the left, 16 subgroups in the center, and among the 5 group-based features in the right.

Figure~\ref{fig:corrg} shows that  the determinant for the individual SHAP is significantly smaller (0.0038), indicating high multicollinearity among the individual features. Grouping features significantly reduces multicollinearity, as evidenced by the increase of the determinant from 0.0038 to 0.0574 for 16 subgroups and further to 0.7522 for 5 groups. This enhances the stability of the model and reduces the risk of overfitting caused by redundant features.



We conclude this section interpreting  what obtained from our analysis to  identifying patterns in financial markets that serve as predictors for the evolution of recovery rates. 
Our findings highlight the critical role of market-related variables and bond characteristics in shaping recovery outcomes, followed by macroeconomic conditions. These findings emphasize the intricate interplay of firm-specific and external economic factors in influencing recovery rates, while the relatively lower importance of industry-specific variables reflects their limited predictive power in this context.

\section{Concluding Remarks }\label{sec:rem}

The paper proposes a novel statistical test to assess the significance of  group Shapley values, thereby enhancing their usability. In comparison with individual Shapley values, group Shapley values can improve both the statistical and the subject matter interpretations of Shapley value explanations.

We illustrate the significance and power of our proposed test by means of a simulation study. And we illustrate the importance of group Shapley values, empowered by our test, in a real problem that concerns the prediction of bond recovery rates. This is a difficult economic problem, for which several possible explanations of the response variable coexist, and have been discussed in the economic and financial literature. We show how group Shapley features can considerably improve the interpretation of the results,in economic terms; but also their statistical properties, such as concentration and correlation.

From an economic viewpoint, the methods proposed in our study can be employed to 
address several crucial  questions, primarily focusing on how feature contributions to recovery rates vary across  financial markets and the roles financial development and regulatory environments play in these variations.

By leveraging Group Shapley values and significance testing, we provide a robust mechanism to pinpoint economically meaningful feature groups, enhancing the interpretability and practical relevance of recovery rate predictions in financial markets. This approach not only helps refine predictive models but also offers actionable insights for risk assessment and management in the evolving dynamics of recovery processes.

The implications of our work for policy and strategy are profound. Policymakers and financial institutions must consider regional economic structures when designing financial strategies and interventions. Addressing disparities in feature contributions can improve economic stability and competitiveness, particularly in emerging markets. These insights emphasize the necessity for tailored financial strategies and enhanced transparency in predictive modeling to manage economic disparities and foster stable financial markets globally.

By integrating Explainable AI techniques like Group SHAP values with disparity assessments, this study provides a nuanced understanding of regional disparities in recovery rates. This approach can inform policies aimed at promoting equitable credit practices, ultimately fostering economic growth, reducing financial risk, and ensuring efficient and just credit systems across different regions.

More generally, the approach presented in this paper, based on group Shapley values, and on the assessment of their statistical significance, can be employed in all types of predictive problems, in economics and finance but also in all types of machine learning applications. Future research shoud consider this type of extension of our proposal.

\newpage

\section*{Appendix: Technical proofs}\label{sec:appendix}
\setcounter{equation}{0} 
\setcounter{subsection}{0}
\renewcommand{\thesubsection}{A.\arabic{subsection}}
\subsection{Lemmas and Proofs}
To derive the asymptotic distribution of $T_{GS}$, we introduce a useful technique called "normal-reference" test. The normal-reference test functions in the following manner:

For the 2 datasets denoted by $\bTheta$ with the corresponding test statistic $T_{1,0}(\bTheta)$, we assume that these follow a normal distribution, represented as $\bTheta$. Then we define the related test statistic $T_{1,0}^*(\bTheta^{*})$ as the "normal-reference test statistic", and its distribution is called the "normal-reference distribution".

\begin{lmm}\label{lmm:Tlim}
    Under Condition C2, we have
\[
\|\calL(\tT_{1,0})-\calL(\tT_{1,0}^*)\|_3\le \frac{(2\gamma)^{3/2}}{3^{1/4}}\Big(\frac{1}{S}\Big)^{1/2},
\]
where $\gamma$ is a constant such that $3 < \gamma < \infty$, and $\|\cdot\|_3$ denotes an order-3 norm of a distribution proposed by \cite{wangxu2021}.
\end{lmm}
\begin{proof}[\textbf{\upshape Proof of Lemma~\ref{lmm:Tlim}}]

    It is worth noting that Lemma~\ref{lmm:Tlim} is largely equivalent to the Theorem 1 in \cite{WANG2024105354}, with the primary difference being in Condition C1 in this study and Condition C2 in \cite{WANG2024105354}, denoted by C2. Therefore, to prove Lemma~\ref{lmm:Tlim} requires demonstrating that Condition C2 will will be satisfied automatically under Condition C1 and zero-mean assumption, which is required in \cite{WANG2024105354}.

    \begin{itemize}
        \item[C2.] There is a universal constant $3\leq\gamma<\infty$ such that for all $p\times K^2$ real matrix $\bB$, we have
$
\E\|\bB\bu_s\|^4\le \gamma \E^2(\|\bB\bu_s\|^2)$ where $1\le p\le K^2$, where $\bu_s=vec(\bphi_s\bphi_s^{T})-\E(vec(\bphi_s\bphi_s^{T}))$ is a centralized induced sample and $vec(\cdot)$ denotes the vectorization operator to stack the column vectors of a matrix one by one. 
    \end{itemize}

     Let $\bB=(\bb_1,\ldots,\bb_p)^T:p\times K^2$. We have $\|\bB\bu_s\|^2=\sum_{\ell=1}^p(\bb_\ell^T\bu_s)^2$. 
     
     For $\ell\in\{1,\ldots,p\}$, we can further write $\bb_\ell^T\bu_s=\bb_\ell^T[\bphi_s\otimes \bphi_s-vec(\bSigma)]=\bphi_s^{T}\bB_\ell^*\bphi_s-\bb_\ell^T vec(\bSigma)$, where $\otimes$ denotes the well-known Kronecker product and 
  \[
 \bB_\ell^*=\begin{pmatrix}
 b_{\ell,1}  &\cdots&b_{\ell,K}\\
 \vdots &\cdots &\vdots\\
 b_{\ell,K^2-K+1}&\cdots&b_{\ell,K^2}
 \end{pmatrix}
  \]
  with $b_{\ell,r},r\in\{1,\ldots,K^2\}$ being the $r$th entry of $\bb_\ell,\ell\in\{1,\ldots,p\}$. 
  
  Since $\bphi_s^{T}\bB_\ell^*\bphi_s$ is a quadratic form,
  it can also be  expressed as $\bphi_s^{T}\bG_{\ell}\bphi_s$
  where $\bG_{\ell}=(\bB_{\ell}^*+\bB_{\ell}^{*^T})/2$ is a symmetric matrix.  Under Condition C2 and the zero-mean condition, we can express $\bphi_s=\bGamma\bz_s$. 
  Then for $\ell\in\{1,\ldots,p\}$, we have $\bb_\ell^T\bu_s=\bz_s^T\bGamma^{(A)^T}\bG_\ell\bGamma\bz_s-\bb_\ell^T vec(\bSigma)=\bz_s^T\bV_{\ell}\bz_s-\tr(\bV_{\ell})$, where $\bV_{\ell}=\bGamma^{(A)^T}\bG_\ell\bGamma$.  
  
  Under Condition C2, by Proposition A1 (iii) of \cite{chenzhangzhong2010JASA}, we have
  \begin{equation}\label{inequ1}
\E(\bb_\ell^T\bu_s)^4=\E[\bz_s^T\bV_{\ell}\bz_s-\tr(\bV_{\ell})]^4\leq C_{\ell}\tr^2(\bV_{\ell}^{2})<\infty,
  \end{equation}
  where $C_{\ell}$ is some positive constant. The above inequality  (\ref{inequ1}) is also independently obtained by \cite[p. 4]{himeno2014estimations} under a similar condition.  
  
  Under Condition C2, by Proposition A1 (i) of \cite{chenzhangzhong2010JASA}, we have $\E(\bz_s^T\bV_{\ell}\bz_s)^2=\tr^2(\bV_{\ell})+2\tr(\bV_{\ell}^{2})+\Delta \tr(\bV_{\ell}o\bV_{\ell})$, where $\Delta$ is given in Condition C2 and $\bA o \bB=(a_{ij}b_{ij})$ when $\bA=(a_{ij})$ and $\bB=(b_{ij})$. It follows that
  \[
  \E(\bb_{\ell}^T \bu_s)^2=\E(\bz_s^T\bV_{\ell}\bz_s)^2-\tr^2(\bV_{\ell})=2\tr(\bV_{\ell}^{2})+\Delta \tr(\bV_{\ell}o\bV_{\ell})\ge \tr(\bV_{\ell}^{2}),
  \]
  provided that $\Delta\ge -1$. Therefore, we have
\begin{equation}\label{inequ2}
\E(\bb_\ell^T\bu_s)^4\leq C_{\ell}\E^2(\bb_\ell^T\bu_s)^2.
\end{equation}  
  By the Cauchy--Schwarz inequality, we have
  \[
\E\left[\sum_{\ell=1}^q (\bb_{\ell}^T \bu_s)^2 \right]^2\le \left[ \sum_{\ell=1}^q \sqrt{\E(\bb_{\ell}^T \bu_s)^4}\right]^2.
  \]
This, together with  (\ref{inequ2}) implies that
\[
\E\|\bB\bu_s\|^4=\E\Big[\sum_{\ell=1}^q(\bb_\ell^T\bu_s)^2\Big]^2\leq\left[\sum_{\ell=1}^q\sqrt{\E(\bb_\ell^T\bu_s)^4}\right]^2\leq\left[\sum_{\ell=1}^q\sqrt{C_{\ell}}\E(\bb_\ell^T\bu_s)^2\right]^2\leq \gamma \E^2\|\bB\bu_s\|^2,
\]
where $\gamma>\max_{\ell=1}^q C_{\ell}$  such that  $3\leq \gamma<\infty$. The proof is complete. 
\end{proof}
    
\begin{proof}[\textbf{\upshape Proof of Theorem~\ref{thm:size}}]
    
    Lemma~\ref{lmm:Tlim} establishes that the distance between the distribution of $\tT_{1,0}$ and its normal-reference counterpart $\tT_{1,0}^*$ is on the order of $\mathcal{O}\{(1/S)^{1/2}\}$. This finding implies that $\calL(\tT_{1,0})$ and $\calL(\tT_{1,0}^*)$ are asymptotically equivalent. Therefore, Lemma~\ref{lmm:Tlim} effectively provides a robust theoretical basis for approximating $\calL(\tT_{1,0})$ using its normal-reference distribution $\calL(\tT_{1,0}^*)$.

    Note that $T_{1,0}^*=\|\barbTheta^{*}\|^2-\tr(\hbSigma)$ can be rewrited as
    \be
    \begin{split}
    \label{T10star}
    T_{1,0}^*&=\frac{2}{S(S-1)}\sum_{1\leq s_i<s_j \leq S} \btheta_{s_i}^{*^T}\btheta_{s_j}^{*} -\frac{2}{S^2}\sum_{s_i=1}^{S} \sum_{s_j=1}^{S} \btheta_{s_i}^{*^T}\btheta_{s_j}^{*}, 
    \end{split}
    \ee
    where $\btheta_s^{*}$ is the normal-reference of the centralized version of $\bphi_s$, denoted by $\btheta_s=\bphi_s-\bmu$. 
    
    From (\ref{T10star}), $T_{1,0}^*$ can be expressed as a quadratic form of normal random variables. Let $\chi_{v}^{2}$ denote a central chi-squared distribution with $v$ degrees of freedom and  $\dequ$ denote equality in distribution, then $T_{1,0}^*$ has the same distribution as that of a chi-squared-type mixture as follows:
    \be\label{T10starchisq}
    T_{1,0}^*\dequ\sum_{r=1}^{K} \lambda_{r}X_r-\left\{\frac{\sum_{r=1}^{K}\lambda_{r} X_{Sr}}{S(S-1)}\right\},
    \ee
    where $\lambda_{r},r\in\{1,\ldots, K\}$ are  the unknown  eigenvalues of $\bSigma$, and
    $X_{r}\iidsim\chi_1^2$ and $X_{S,r} \iidsim\chi_{S-1}^2$ are mutually independent.
     
    Then, by Theorem 1 (b) of \cite{Zhang2005}, $T_{1,0}^*$ (\ref{T10starchisq}) will converge to $R$ (\ref{R}).

    Since under the null hypothesis (\ref{jointtest}), as $S,K\to\infty$, $P(T_0|H_0)\to 0$ (Proposition 4.1 in \cite{fan2015power}). Therefore,
    \[
        T_{GS} = T_0 + T_1 \to R.
    \]
    
\end{proof}

\subsection{Estimators}

It requires consistent estimators of $\tr(\bSigma)$, $\tr(\bSigma^{2})$ and $\tr(\bSigma^{3})$ to obtain the estimated $\widehat{\calK_2(T_{1,0})}$ and $\widehat{\calK_3(T_{1,0})}$. Recall the usual unbiased estimator of $\bSigma$ is given by $\hbSigma$ (\ref{hbSigma}). By Lemma S3. in \cite{zhangguozhoucheng2020JASA}, the estimator of $\tr(\bSigma^{2})$ is given by
$\widehat{\tr(\bSigma^{2})}=(S-1)^2\{(S-2)(S+1)\}^{-1}\{\tr(\bSigma^{2})-\tr^2(\bSigma)/(S-1)\}$. Based on (\ref{K2T10}), the unbiased estimated $\calK_2(T_{1,0})$ can be formulated as
\be\label{hK2T10}
\widehat{\calK_2(T_{1,0})}=2\left\{\frac{\widehat{\tr(\bSigma^{2})}}{S(S-1)}\right\}.
\ee
By Lemma 1 of \cite{zhangzhouguo2020onesamp}, the estimator of $\tr(\bSigma^{3})$ is given by 
\begin{align*}
    \widehat{\tr(\bSigma^{3})}=&(S-1)^4\{(S^2+S-6)(S^2-2S-3)\}^{-1}\\
    &\{\tr(\hbSigma^{3})-3(S-1)^{-1}\tr(\hbSigma)\tr(\hbSigma^{2})
+2(S-1)^{-2}\tr^3(\hbSigma)\}.
\end{align*}

From (\ref{K3T10}), the unbiased estimator of $\calK_3(T_{1,0})$ can be computed by
\be\label{hK3T10}
\widehat{\calK_3(T_{1,0})}= 8\left\{\frac{(S-2)\widehat{\tr(\bSigma^{3})}}{S^2(S-1)^2}\right\}.
\ee
Following from (\ref{betadf}), (\ref{hK2T10}) and (\ref{hK3T10}), it leads to
\be\label{hbetadf}
    \hbeta_0=-\frac{2\widehat{\calK_2^2(T_{1,0})}}{\widehat{\calK_3(T_{1,0})}},\;\;
\hbeta_1=\frac{\widehat{\calK_3(T_{1,0})}}{4\widehat{\calK_2(T_{1,0})}},\;\;\mbox{and}\;\;
\hd=\frac{8\widehat{\calK_2^3(T_{1,0})}}{\widehat{\calK_3^2(T_{1,0})}}.
\ee

\newpage

\bibliographystyle{apalike}
\bibliography{GS}

\begin{thebibliography}{}

\bibitem[Ahad et~al., 2011]{ahad2011sensitivity}
Ahad, N.~A., Yin, T.~S., Othman, A.~R., and Yaacob, C.~R. (2011).
\newblock Sensitivity of normality tests to non-normal data.
\newblock {\em Sains Malaysiana}, 40(6):637--641.

\bibitem[Altman et~al., 2005]{altman2005link}
Altman, E.~I., Brady, B., Resti, A., and Sironi, A. (2005).
\newblock The link between default and recovery rates: Theory, empirical evidence, and implications.
\newblock {\em The Journal of Business}, 78(6):2203--2228.

\bibitem[Altman and Kishore, 1996]{altman1996almost}
Altman, E.~I. and Kishore, V.~M. (1996).
\newblock Almost everything you wanted to know about recoveries on defaulted bonds.
\newblock {\em Financial Analysts Journal}, 52(6):57--64.

\bibitem[Aoshima and Yata, 2018]{aoshima2018two}
Aoshima, M. and Yata, K. (2018).
\newblock Two-sample tests for high-dimension, strongly spiked eigenvalue models.
\newblock {\em Statistica Sinica}, pages 43--62.

\bibitem[Bai and Saranadasa, 1996]{BS1996}
Bai, Z. and Saranadasa, H. (1996).
\newblock Effect of high dimension: By an example of a two sample problem.
\newblock {\em Statistica Sinica}, 6(2):311--329.

\bibitem[Bellotti et~al., 2021]{BELLOTTI2021428}
Bellotti, A., Brigo, D., Gambetti, P., and Vrins, F. (2021).
\newblock Forecasting recovery rates on non-performing loans with machine learning.
\newblock {\em International Journal of Forecasting}, 37(1):428--444.

\bibitem[Bruche and Gonz{\'a}lez-Aguado, 2010]{Bruche2010rrpd}
Bruche, M. and Gonz{\'a}lez-Aguado, C. (2010).
\newblock Recovery rates, default probabilities, and the credit cycle.
\newblock {\em Journal of Banking \& Finance}, 34(4):754--764.
\newblock INTERACTION OF MARKET AND CREDIT RISK.

\bibitem[Cantor et~al., 2008]{cantor2008default}
Cantor, R., Emery, F., Kim, K., Ou, S., and Tennant, J. (2008).
\newblock Default and recovery rates of corporate bond issuers.
\newblock {\em Moody's Investors Service, Global Credit Research}.

\bibitem[Chen and Qin, 2010]{chen2010two}
Chen, S.~X. and Qin, Y.-L. (2010).
\newblock {A two-sample test for high-dimensional data with applications to gene-set testing}.
\newblock {\em The Annals of Statistics}, 38(2):808 -- 835.

\bibitem[Chen et~al., 2010]{chenzhangzhong2010JASA}
Chen, S.~X., Zhang, L.-X., and Zhong, P.-S. (2010).
\newblock Tests for high-dimensional covariance matrices.
\newblock {\em Journal of the American Statistical Association}, 105(490):810--819.

\bibitem[Cochrane and Piazzesi, 2005]{cochrane2005bond}
Cochrane, J.~H. and Piazzesi, M. (2005).
\newblock Bond risk premia.
\newblock {\em American economic review}, 95(1):138--160.

\bibitem[Das and Hanouna, 2009]{das2009implied}
Das, S.~R. and Hanouna, P. (2009).
\newblock Implied recovery.
\newblock {\em Journal of Economic Dynamics and Control}, 33(11):1837--1857.

\bibitem[Datta et~al., 2016]{datta2016algorithmic}
Datta, A., Sen, S., and Zick, Y. (2016).
\newblock Algorithmic transparency via quantitative input influence: Theory and experiments with learning systems.
\newblock In {\em 2016 IEEE symposium on security and privacy (SP)}, pages 598--617. IEEE.

\bibitem[Efron, 1981]{efron1981nonparametric}
Efron, B. (1981).
\newblock Nonparametric standard errors and confidence intervals.
\newblock {\em canadian Journal of Statistics}, 9(2):139--158.

\bibitem[Fan et~al., 2015]{fan2015power}
Fan, J., Liao, Y., and Yao, J. (2015).
\newblock Power enhancement in high-dimensional cross-sectional tests.
\newblock {\em Econometrica}, 83(4):1497--1541.

\bibitem[Fryer et~al., 2020]{fryer2020shapley}
Fryer, D., Str{\"u}mke, I., and Nguyen, H. (2020).
\newblock Shapley value confidence intervals for attributing variance explained.
\newblock {\em Frontiers in Applied Mathematics and Statistics}, 6:587199.

\bibitem[Geurts et~al., 2006]{geurts2006extremely}
Geurts, P., Ernst, D., and Wehenkel, L. (2006).
\newblock Extremely randomized trees.
\newblock {\em Machine learning}, 63:3--42.

\bibitem[Gompers and Lerner, 1998]{slrun1998}
Gompers, P. and Lerner, J. (1998).
\newblock Venture capital distributions: Short-run and long-run reactions.
\newblock {\em The Journal of Finance}, 53(6):2161--2183.

\bibitem[Hamilton and Papadopoulos, 2023]{hamilton2023using}
Hamilton, R.~I. and Papadopoulos, P.~N. (2023).
\newblock Using shap values and machine learning to understand trends in the transient stability limit.
\newblock {\em IEEE Transactions on Power Systems}.

\bibitem[Himeno and Yamada, 2014]{himeno2014estimations}
Himeno, T. and Yamada, T. (2014).
\newblock Estimations for some functions of covariance matrix in high dimension under non-normality and its applications.
\newblock {\em Journal of Multivariate Analysis}, 130:27--44.

\bibitem[Huang and Huang, 2023]{huang2023increasing}
Huang, A.~A. and Huang, S.~Y. (2023).
\newblock Increasing transparency in machine learning through bootstrap simulation and shapely additive explanations.
\newblock {\em PLoS One}, 18(2):e0281922.

\bibitem[Jankowitsch et~al., 2014a]{jankowitsch2014determinants}
Jankowitsch, R., Nagler, F., and Subrahmanyam, M.~G. (2014a).
\newblock The determinants of recovery rates in the us corporate bond market.
\newblock {\em Journal of Financial Economics}, 114(1):155--177.

\bibitem[Jankowitsch et~al., 2014b]{JR2014determinant}
Jankowitsch, R., Nagler, F., and Subrahmanyam, M.~G. (2014b).
\newblock The determinants of recovery rates in the us corporate bond market.
\newblock {\em Journal of Financial Economics}, 114(1):155--177.

\bibitem[Janssen et~al., 2022]{janssen2022application}
Janssen, A., Hoogendoorn, M., Cnossen, M.~H., Math{\^o}t, R.~A., Group, O.-C.~S., Consortium, S., Cnossen, M., Reitsma, S., Leebeek, F., Math{\^o}t, R., Fijnvandraat, K., et~al. (2022).
\newblock Application of shap values for inferring the optimal functional form of covariates in pharmacokinetic modeling.
\newblock {\em CPT: Pharmacometrics \& Systems Pharmacology}, 11(8):1100--1110.

\bibitem[Jullum et~al., 2021]{jullum2021groupshapley}
Jullum, M., Redelmeier, A., and Aas, K. (2021).
\newblock groupshapley: Efficient prediction explanation with shapley values for feature groups.

\bibitem[Katayama et~al., 2013]{Katayama2013}
Katayama, S., Kano, Y., and Srivastava, M.~S. (2013).
\newblock Asymptotic distributions of some test criteria for the mean vector with fewer observations than the dimension.
\newblock {\em Journal of Multivariate Analysis}, 116:410--421.

\bibitem[Lipsitz et~al., 1998]{lipsitz1998tests}
Lipsitz, S.~R., Dear, K.~B., Laird, N.~M., and Molenberghs, G. (1998).
\newblock Tests for homogeneity of the risk difference when data are sparse.
\newblock {\em Biometrics}, pages 148--160.

\bibitem[Lundberg et~al., 2020]{TreeSHAP2020}
Lundberg, S.~M., Erion, G., Chen, H., DeGrave, A., Prutkin, J.~M., Nair, B., Katz, R., Himmelfarb, J., Bansal, N., and Lee, S.-I. (2020).
\newblock From local explanations to global understanding with explainable ai for trees.
\newblock {\em Nature machine intelligence}, 2(1):56--67.

\bibitem[Lundberg and Lee, 2017a]{SHAP2017}
Lundberg, S.~M. and Lee, S.-I. (2017a).
\newblock A unified approach to interpreting model predictions.
\newblock In Guyon, I., Luxburg, U.~V., Bengio, S., Wallach, H., Fergus, R., Vishwanathan, S., and Garnett, R., editors, {\em Advances in Neural Information Processing Systems}, volume~30. Curran Associates, Inc.

\bibitem[Lundberg and Lee, 2017b]{lundberg2017unified}
Lundberg, S.~M. and Lee, S.-I. (2017b).
\newblock A unified approach to interpreting model predictions.
\newblock {\em Advances in neural information processing systems}, 30.

\bibitem[Menze et~al., 2011]{menze2011oblique}
Menze, B.~H., Kelm, B.~M., Splitthoff, D.~N., Koethe, U., and Hamprecht, F.~A. (2011).
\newblock On oblique random forests.
\newblock In {\em Machine Learning and Knowledge Discovery in Databases: European Conference, ECML PKDD 2011, Athens, Greece, September 5-9, 2011, Proceedings, Part II 22}, pages 453--469. Springer.

\bibitem[Mora, 2006]{mora2006sovereign}
Mora, N. (2006).
\newblock Sovereign credit ratings: Guilty beyond reasonable doubt?
\newblock {\em Journal of Banking \& Finance}, 30(7):2041--2062.

\bibitem[Moretti et~al., 2008]{moretti2008combining}
Moretti, S., van Leeuwen, D., Gmuender, H., Bonassi, S., Van~Delft, J., Kleinjans, J., Patrone, F., and Merlo, D.~F. (2008).
\newblock Combining shapley value and statistics to the analysis of gene expression data in children exposed to air pollution.
\newblock {\em BMC bioinformatics}, 9:1--21.

\bibitem[Murthy et~al., 1994]{murthy1994system}
Murthy, S.~K., Kasif, S., and Salzberg, S. (1994).
\newblock A system for induction of oblique decision trees.
\newblock {\em Journal of artificial intelligence research}, 2:1--32.

\bibitem[Nazemi et~al., 2018]{NAZEMI2018664}
Nazemi, A., Heidenreich, K., and Fabozzi, F.~J. (2018).
\newblock Improving corporate bond recovery rate prediction using multi-factor support vector regressions.
\newblock {\em European Journal of Operational Research}, 271(2):664--675.

\bibitem[Pettit and Venkatesh, 1995]{longrun1995}
Pettit, R.~R. and Venkatesh, P.~C. (1995).
\newblock Insider trading and long-run return performance.
\newblock {\em Financial Management}, 24(2):88--103.

\bibitem[Roder et~al., 2021]{roder2021explaining}
Roder, J., Maguire, L., Georgantas, R., and Roder, H. (2021).
\newblock Explaining multivariate molecular diagnostic tests via shapley values.
\newblock {\em BMC Medical Informatics and Decision Making}, 21:1--18.

\bibitem[Strumbelj and Kononenko, 2010]{strumbelj2010efficient}
Strumbelj, E. and Kononenko, I. (2010).
\newblock An efficient explanation of individual classifications using game theory.
\newblock {\em The Journal of Machine Learning Research}, 11:1--18.

\bibitem[Strumbelj and Kononenko, 2014]{vstrumbelj2014explaining}
Strumbelj, E. and Kononenko, I. (2014).
\newblock Explaining prediction models and individual predictions with feature contributions.
\newblock {\em Knowledge and information systems}, 41:647--665.

\bibitem[Wang et~al., 2024]{WANG2024105354}
Wang, J., Zhu, T., and Zhang, J.-T. (2024).
\newblock Two-sample test for high-dimensional covariance matrices: A normal-reference approach.
\newblock {\em Journal of Multivariate Analysis}, 204:105354.

\bibitem[Wang and Xu, 2022]{wangxu2021}
Wang, R. and Xu, W. (2022).
\newblock {An approximate randomization test for the high-dimensional two-sample {Behrens--Fisher} problem under arbitrary covariances}.
\newblock {\em Biometrika}, 109(4):1117--1132.

\bibitem[Yu et~al., 2024]{yu2024power}
Yu, X., Yao, J., and Xue, L. (2024).
\newblock Power enhancement for testing multi-factor asset pricing models via fisher's method.
\newblock {\em Journal of Econometrics}, 239(2):105458.

\bibitem[Zhang, 2005]{Zhang2005}
Zhang, J.-T. (2005).
\newblock Approximate and asymptotic distributions of chi-squared-type mixtures with applications.
\newblock {\em Journal of the American Statistical Association}, 100(469):273--285.

\bibitem[Zhang et~al., 2020]{zhangguozhoucheng2020JASA}
Zhang, J.-T., Guo, J., Zhou, B., and Cheng, M.-Y. (2020).
\newblock A simple two-sample test in high dimensions based on {$L^2$}-norm.
\newblock {\em Journal of American Statistical Association}, 115(530):1011--1027.

\bibitem[Zhang et~al., 2022]{zhangzhouguo2020onesamp}
Zhang, J.-T., Zhou, B., and Guo, J. (2022).
\newblock Testing high-dimensional mean vector with applications.
\newblock {\em Statistical Papers}, 63(4):1105--1137.

\bibitem[Zhang et~al., 2021]{zhangguozhouzhu2020}
Zhang, J.-T., Zhou, B., Guo, J., and Zhu, T. (2021).
\newblock Two-sample {Behrens--Fisher} problems for high-dimensional data: A normal reference approach.
\newblock {\em Journal of Statistical Planning and Inference}, 213:142--161.

\end{thebibliography}
\end{document}